\documentclass{article} 
\PassOptionsToPackage{table}{xcolor} 
\usepackage{iclr2027_conference,times}

\usepackage{amsmath,amsfonts,bm}

\def\eqref#1{equation~\ref{#1}}

\def\1{\bm{1}}

\DeclareMathAlphabet{\mathsfit}{\encodingdefault}{\sfdefault}{m}{sl}
\SetMathAlphabet{\mathsfit}{bold}{\encodingdefault}{\sfdefault}{bx}{n}

\usepackage[hidelinks]{hyperref}
\usepackage{url}
\usepackage{tikz}
\usepackage{soul}
\usepackage[capitalize]{cleveref}
\usepackage{booktabs}
\usepackage{subcaption}
\usepackage{multirow}
\usepackage{graphicx}
\usepackage{algorithm}
\usepackage{algpseudocode}
\usepackage{wrapfig}
\usepackage{capt-of}
\usepackage{placeins}

\newcommand{\dataset}{RoboQuest}
\newcommand{\gptsfull}{GPT-6 Astra}
\newcommand{\opusfull}{Claude Opus 5.5}
\newcommand{\solfull}{GPT-6.1 Sol}
\newcommand{\fablefull}{Claude Fable 5.1}
\newcommand{\geminifull}{Gemini 3.8 Flash}
\newcommand{\argonfull}{Gemini 4 Argon}
\newcommand{\argon}{Gemini 4 Argon}
\newcommand{\gpts}{GPT-6 Astra}
\newcommand{\opus}{Opus 5.5}
\newcommand{\sol}{GPT-6.1 Sol}
\newcommand{\fable}{Fable 5.1}
\newcommand{\gemini}{Gemini 3.8 Flash}
\definecolor{rqmainheat}{HTML}{BE7937}
\colorlet{mainheat}{rqmainheat!55!white}
\newcommand{\mh}[1]{\cellcolor{mainheat!#1!white}#1}
\newcommand{\mhb}[1]{\cellcolor{mainheat!#1!white}\textbf{#1}}
\definecolor{skillok}{HTML}{E3F4E6}
\definecolor{skillbad}{HTML}{F9E3E0}
\newcommand{\sk}[1]{\ifnum#1<60 \cellcolor{skillbad}\else\ifnum#1>89 \cellcolor{skillok}\fi\fi #1}
\newcommand{\skb}[1]{\ifnum#1<60 \cellcolor{skillbad}\else\ifnum#1>89 \cellcolor{skillok}\fi\fi \textbf{#1}}

\usepackage{booktabs}
\usepackage{tabularx}
\usepackage{array}
\usepackage[table]{xcolor}
\usepackage{amssymb}
\usepackage{makecell}
\usepackage{colortbl}
\usepackage{pgf}

\definecolor{RQOrange}{HTML}{D97706}
\definecolor{RQHighlight}{HTML}{FFF5E8}
\definecolor{RQGreen}{HTML}{208060}
\definecolor{RQAmber}{HTML}{C88719}
\definecolor{RQGray}{HTML}{8A969D}
\definecolor{RQDark}{HTML}{24313A}

\newcommand{\no}{\textcolor{RQGray}{--}}

\let\realcite\citep
\renewcommand{\cite}[1]{\ifx.#1.\hl{[?]}\else\realcite{#1}\fi}

\crefformat{section}{\S#2#1#3} 
\crefformat{subsection}{\S#2#1#3}
\crefformat{subsubsection}{\S#2#1#3}

\newcolumntype{C}{>{\centering\arraybackslash}p{1.05cm}}
\newcolumntype{M}{>{\centering\arraybackslash}p{1.25cm}}
\newcolumntype{D}{>{\centering\arraybackslash}p{1.75cm}}
\newcolumntype{P}{>{\raggedright\arraybackslash}X}

\newcolumntype{C}{>{\centering\arraybackslash}p{1.05cm}}
\newcolumntype{M}{>{\centering\arraybackslash}p{1.25cm}}
\newcolumntype{D}{>{\centering\arraybackslash}p{1.75cm}}
\newcolumntype{P}{>{\raggedright\arraybackslash}X}

\newcolumntype{L}[1]{>{\raggedright\arraybackslash}m{#1}}
\newcolumntype{C}[1]{>{\centering\arraybackslash}m{#1}}
\newcolumntype{Y}{>{\raggedright\arraybackslash}X}

\title{\dataset{}: Generalist Physical Agents that Search, Inspect and Test}

\author{Renhang Liu$^{1}$\quad Navonil Majumder$^{1}$\quad Tej Deep Pala$^{1}$\quad Soujanya Poria$^{1}$\\
\normalfont $^{1}$Nanyang Technological University\\[10pt]
\normalfont \textbf{Project page:} \url{https://declare-lab.github.io/RoboQuest/}}

\iclrfinalcopy 
\begin{document}
\raggedbottom
\addtocontents{toc}{\protect\setcounter{tocdepth}{-1}}

\maketitle
\fancyhead{}\renewcommand{\headrulewidth}{0pt}
\suppressfloats[t]

\begin{abstract}
Recent advances in multimodal foundation models have made them capable generalist physical agents for a range of manipulation tasks. However, successful operation in an unfamiliar environment may require an agent to seek task-relevant information through interaction when it is absent from the observations: it may need to determine where a relevant object is, inspect an unobserved property, or discover the effect of an unfamiliar tool. We thus introduce \dataset{}, a benchmark for goal-directed embodied exploration, where agents must actively acquire task-relevant information through physical interaction, use the resulting evidence to adapt subsequent actions, and autonomously decide when to commit to task completion. \dataset{} comprises ten mobile manipulation tasks centered on three forms of uncertainty: search, manipulation-based inspection, and interactive testing. We evaluate six frontier multimodal agents and two smaller open models through a common visuomotor interface, as well as a $\pi_{0.5}$ policy fine-tuned on the full-episode demonstrations we release. The best agent succeeds in only 24\% of the episodes, and the fine-tuned policy almost never succeeds. Isolated tests of the execution skills the tasks are built from, with the hidden information supplied, show that the agents can carry out most of the required actions, and our failure analysis finds that execution is not the main bottleneck. Our failure analysis further finds that the agents often stop exploring too early as they make decisions before observing the required evidence for task completion. We also find that agents rarely prevent or repair the disturbances caused by their exploration. Moreover, learning by trial and error remains difficult for most models.
\end{abstract}

\section{Introduction}
\begin{figure}[t]
    \centering
    \includegraphics[width=0.95\linewidth]{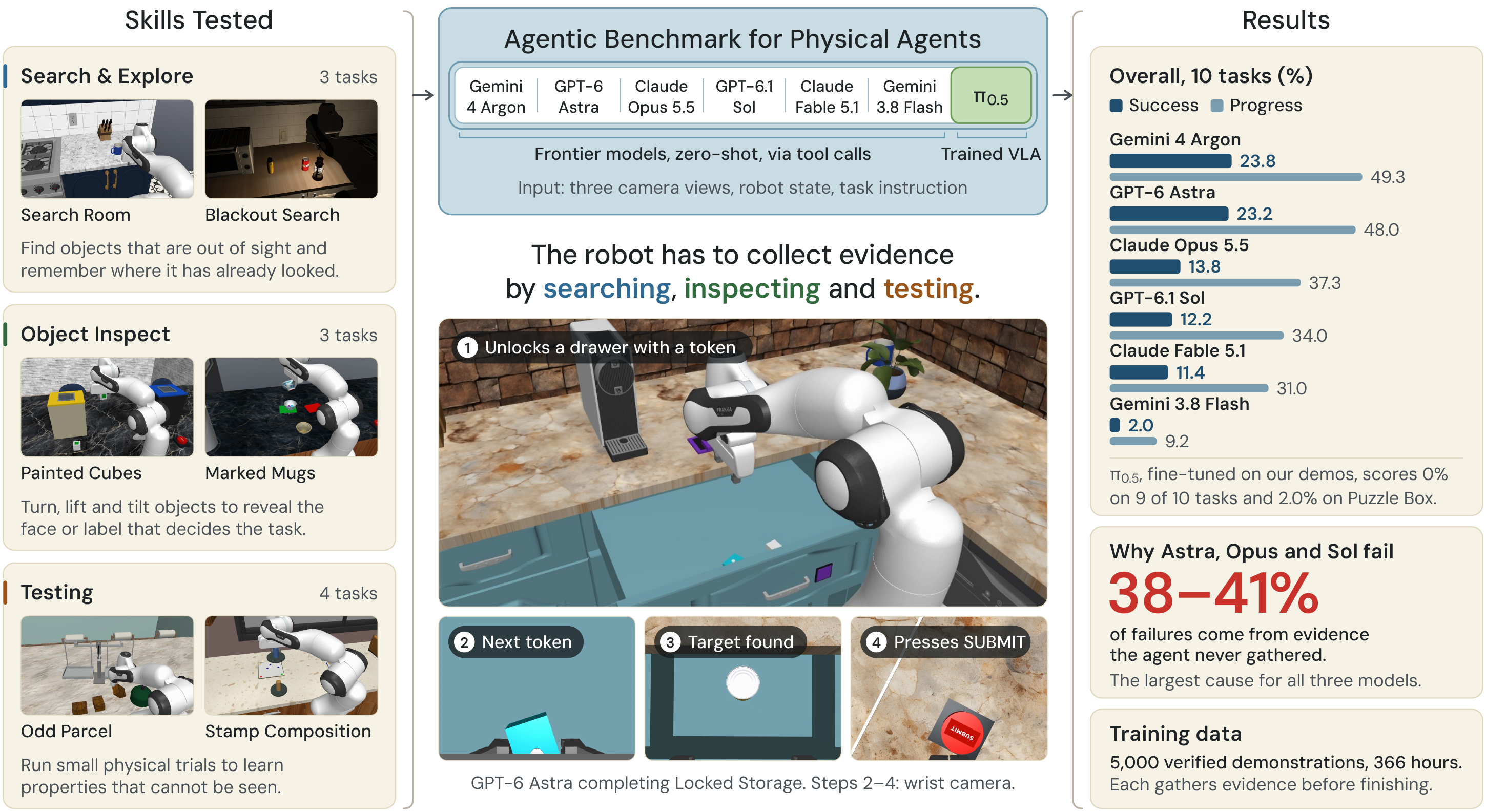}
    \caption{A visual depiction of the salient features of \dataset{} that includes tasks demanding \emph{information-seeking physical interactions} with active perception.}
    \label{fig:teaser}
\end{figure}
Recent advances in generalist robotic agents have enabled increasingly complex manipulation, yet such strong execution may not necessarily translate into successful task completion without ready access to key task-relevant information in the observation. In such settings, the robot must actively determine what it needs to know, how to obtain that information, and how each new observation should shape what it does next. This may require searching an environment, inspecting objects from informative configurations, or deliberately interacting with the world to reveal its behavior. Current robotic manipulation benchmarks~\cite{robocasa365,chen2026rmbench,tao2024maniskill3,li2026libero} do not demand such open-ended information-seeking physical interactions from the agents and policies. With the increasing preponderance of generalist robotic agents, such exploratory benchmarking becomes pivotal to truly evaluate the zero-shot capabilities of such agents that may encounter unseen challenges in new environments.

Consider a situation where a robot is searching an unfamiliar room in darkness with only a portable light source. The robot can observe only what the light is aimed at. The challenge here comes from the dual use of the same gripper to carry the light and retrieve objects as well. Exploration is also consequential; for example, a searched compartment may contain the target, nothing, or a key that can be used to access other regions. In other words, each observation can change both its beliefs and next possible actions. The robot must therefore decide where to look, how each discovery should redirect its behavior, and when it has learned enough to act. Similar challenges arise beyond search when a robot must determine whether an object has been sufficiently inspected or choose a physical test to probe certain hidden property of an object.

The underlying principle is longstanding. Active and interactive perception study acquiring task-relevant information through decisive actions rather than passively consuming fixed observations \cite{5968,bohg2017interactive}. However, evaluating this capability requires more than simply hiding information from the initial observation. If every uncertainty has an obvious one-step resolution, successful completion primarily tests if the agent performs that prescribed interaction. Thus, the challenge is beyond simply unveiling missing information, that is autonomously conducting an effective investigation in service of a physical goal, where different actions may expose different evidence. Furthermore, information may need to be accumulated across interactions and acquiring them may affect the state of the task. This question is becoming increasingly palpable as frontier multimodal agents such as \gptsfull{} can now act directly as robot policies across diverse and previously unseen manipulation tasks without task-specific fine-tuning. Improved execution shifts the bottleneck toward deciding what to investigate, how to investigate it, and when enough has been learned.

Thus, we introduce \textbf{RoboQuest}, a benchmark for assessing \emph{goal-directed embodied exploration}, in which a robot must autonomously gather task-relevant evidence through physical interaction, use that evidence to guide subsequent actions, and decide when it knows enough to commit to a solution. RoboQuest contains ten tasks organized into three families: \textbf{Search \& Explore}, \textbf{Object Inspect}, and \textbf{Testing}. These are mobile manipulation tasks set in full kitchens with a mobile manipulator: exploring often means traversing the kitchen as well as precisely manipulating objects. The tasks involve uncertainty about the location and properties of the relevant objects, and making inferences from the outcomes of physical interactions. Rather than instructing the robot with the investigation approach, each task just specifies an externally verifiable physical goal and leaves the investigation strategy to the agent---from evidence gathering steps to conditions of evidence sufficiency and commitment.

Our evaluations reveal a substantial gap between general manipulation competence and the ability to explore effectively under uncertainty. The best frontier agent succeeds in only 24\% of the episodes, and the fine-tuned $\pi_{0.5}$ seldom succeeds. Yet the strongest agents execute most of the required physical actions in isolation when the hidden information is supplied (\cref{sec:control}), and our failure analysis (\cref{sec:analysis}) finds that execution is not the main bottleneck. Most failures instead involve exploration and decision making. The agents often stop exploring too early, or prematurely make decisions before observing their ramifications. Their exploration also disturbs the scenes that they rarely prevent or repair. Learning by trial and error remains difficult for most models. These results highlight a persistent gap between stronger robotic execution and the ability to autonomously explore, resolve uncertainty, and achieve a given physical goal.

Our main contributions are:
\begin{itemize}
    \item \textbf{RoboQuest}, a benchmark for goal-directed embodied exploration with ten carefully designed tasks in which exploration shapes subsequent decisions and final task success.
    \item \textbf{A demonstration dataset that explores}, with 5{,}000 episodes (500 per task, 366 hours at 20\,Hz) from scripts that can easily generate more. The demonstrations show how a robot explores and then acts, with a subtask caption on every frame.
    \item \textbf{A systematic evaluation of general-purpose and trained robotic agents}, including six frontier multimodal agents, two smaller open models and a RoboQuest-finetuned $\pi_{0.5}$, with isolated tests of the execution skills the tasks require and a detailed failure analysis, which together characterize where exploration under uncertainty breaks down.
\end{itemize}

\section{Related Work}
\label{sec:related-work}

\paragraph{Robotic Manipulation Benchmarks.}
Standard manipulation benchmarks such as RLBench~\citep{james2020rlbench}, CALVIN~\citep{mees2022calvin}, LIBERO~\citep{liu2023libero}, ManiSkill3~\citep{tao2024maniskill3}, RoboTwin~2.0~\citep{chen2025robotwin}, BEHAVIOR-1K~\citep{li2023behavior}, and RoboCasa~\citep{nasiriany2024robocasa,robocasa365} evaluate multi-task learning and long-horizon motor control. However, these benchmarks operate under complete or immediately accessible task observability, making the objective primarily execution-centric (\cref{tab:benchmark_comparison}). In contrast, \dataset{} focuses on goal-directed embodied exploration: task-critical information is fundamentally absent from initial observations and cannot be resolved by passive viewing, requiring physical interaction to gather evidence before goal completion.

\paragraph{Comparison with other benchmarks for physical agents.} In most manipulation benchmarks, relevant information for task progression and completion is always present in the current observation/state, as compared in \cref{tab:benchmark_comparison}. In RLBench, LIBERO, and CALVIN, the agent must exercise the correct motor skills, but need not explore to uncover some necessary information. In contrast, every \dataset{} task requires information acquisition. Moreover, success still depends on perception and motor execution as well. Aggregate success rates therefore cannot isolate and probe a single capability. Hence, we analyze every episode in full to examine what evidence the agent gathered, retained, and used before it committed to some action (\cref{sec:analysis}).

\begin{table}[h!]
\centering
\caption{\textbf{Comparison with representative manipulation benchmarks.}
$\checkmark$: explicit target; $\sim$: partial; --: not targeted.
\emph{Hidden}: task information absent from the current observation.}
\label{tab:benchmark_comparison}

\footnotesize
\setlength{\tabcolsep}{4pt}
\resizebox{\textwidth}{!}{%
\begin{tabular}{@{}llccccccc@{}}
\toprule
& & \multicolumn{4}{c}{\textbf{Benchmark features}}
& \multicolumn{3}{c}{\textbf{Evidence seeking}} \\
\cmidrule(lr){3-6}
\cmidrule(l){7-9}

\textbf{Benchmark} & \textbf{Focus}
& \textbf{\makecell{Long\\horizon}}
& \textbf{Hidden} & \textbf{Memory} & \textbf{Mobile}
& \textbf{Search} & \textbf{Inspect} & \textbf{Test} \\
\midrule

RLBench~\citep{james2020rlbench}
& Visuomotor skills
& $\sim$ & $\sim$ & -- & --
& -- & -- & -- \\

ManiSkill3~\citep{tao2024maniskill3}
& Scalable manipulation
& $\sim$ & -- & -- & --
& -- & -- & -- \\

RoboTwin 2.0~\citep{chen2025robotwin}
& Bimanual manipulation
& $\sim$ & -- & -- & --
& -- & $\sim$ & -- \\

LIBERO~\citep{liu2023libero}
& Lifelong skill transfer
& $\checkmark$ & -- & -- & --
& -- & -- & -- \\

CALVIN~\citep{mees2022calvin}
& Long-horizon manipulation
& $\checkmark$ & $\sim$ & -- & --
& -- & -- & -- \\

BEHAVIOR-1K~\citep{li2023behavior}
& Household activities
& $\checkmark$ & $\sim$ & $\sim$ & $\checkmark$
& $\checkmark$ & -- & -- \\

RoboCasa365~\cite{robocasa365}
& Manipulation in kitchen
& $\checkmark$ & -- & $\checkmark$ & $\checkmark$
& $\sim$ & $\sim$ & $\sim$ \\

RMBench~\citep{chen2026rmbench}
& Memory-dependent manipulation
& $\checkmark$ & $\checkmark$ & $\checkmark$ & --
& -- & -- & $\checkmark$ \\

\midrule
\rowcolor{RQHighlight}
\textbf{\textcolor{RQOrange}{RoboQuest}}
& \textbf{Evidence-seeking manipulation}
& $\checkmark$ & $\checkmark$ & $\checkmark$ & $\checkmark$
& $\checkmark$ & $\checkmark$ & $\checkmark$ \\

\bottomrule
\end{tabular}%
}
\end{table}

\paragraph{Active Perception and Interactive Exploration.}
When task relevant information is not immediately observable, robots must rely on active perception to acquire missing information~\citep{bohg2017interactive}. While recent benchmarks investigate camera repositioning~\citep{liu2026activevla,liu2026sapave}, visual occluders~\citep{li2026libero}, container retrieval~\citep{he2026towards,li2026act}, or history retention~\citep{chen2026rmbench}, they restrict active sensing to viewpoint adjustment or single-step un-occlusion. \dataset{} expands physical exploration into three dimensions: directed search, active inspection, and interactive testing. These exploratory interactions can alter the environment in irreversible ways, forcing agents to reason about the consequences of their actions.

\paragraph{Embodied Reasoning and VLAs under Uncertainty.}
Robotic foundation models either employ frontier multimodal models as zero-shot high-level planners~\citep{liang2023code,yang2025embodiedbench} or train Vision-Language-Action (VLA) models for direct low-level control~\citep{brohan2023rt2,kim2025fine,intelligence2025pi}. Existing evaluations test both paradigms in fully observed scenes. When information must be actively gathered, reasoning agents often prematurely commit or hallucinate unobserved states, while Markovian VLAs struggle with multi-step hypothesis testing. \dataset{} provides an architecture-neutral benchmark to evaluate both paradigms under physical uncertainty.

\section{\dataset{} Benchmark}
\label{sec:dataset}

\dataset{} is a collection of ten tasks---detailed in \autoref{sec:tasks}---requiring information-seeking physical interactions for successful completion. As visualized in \cref{fig:teaser} the information seeking interactions constitute three dominant types: search, inspect, and test. In \emph{Search} tasks, the missing information is the location of some object. In \emph{Inspect} tasks, it is some hidden information about an object, and the agent reveals it by interacting with the object: turning, lifting, or opening it. In \emph{Test} tasks, the agent cannot observe the hidden information directly. Thus, it must run a series of trials and identify the relevant cue from the outcomes: an impression, a balance, a rolling ball, or a part that moves. We build the tasks from basic manipulation skills, such as picking and placing, opening a drawer or turning a cube. Some of these skills remain demanding, but more difficulty comes from deciding what to search, inspect or test, and when enough is known (\cref{sec:analysis}). We discuss the capabilities required to carry out these tasks in \cref{sec:capabilities} and from that lens we also analyze the potential failure modes in \cref{sec:analysis}.

\subsection{Benchmark Setup}
\label{sec:setup}

\begin{figure}[t]
\centering
\includegraphics[width=\textwidth]{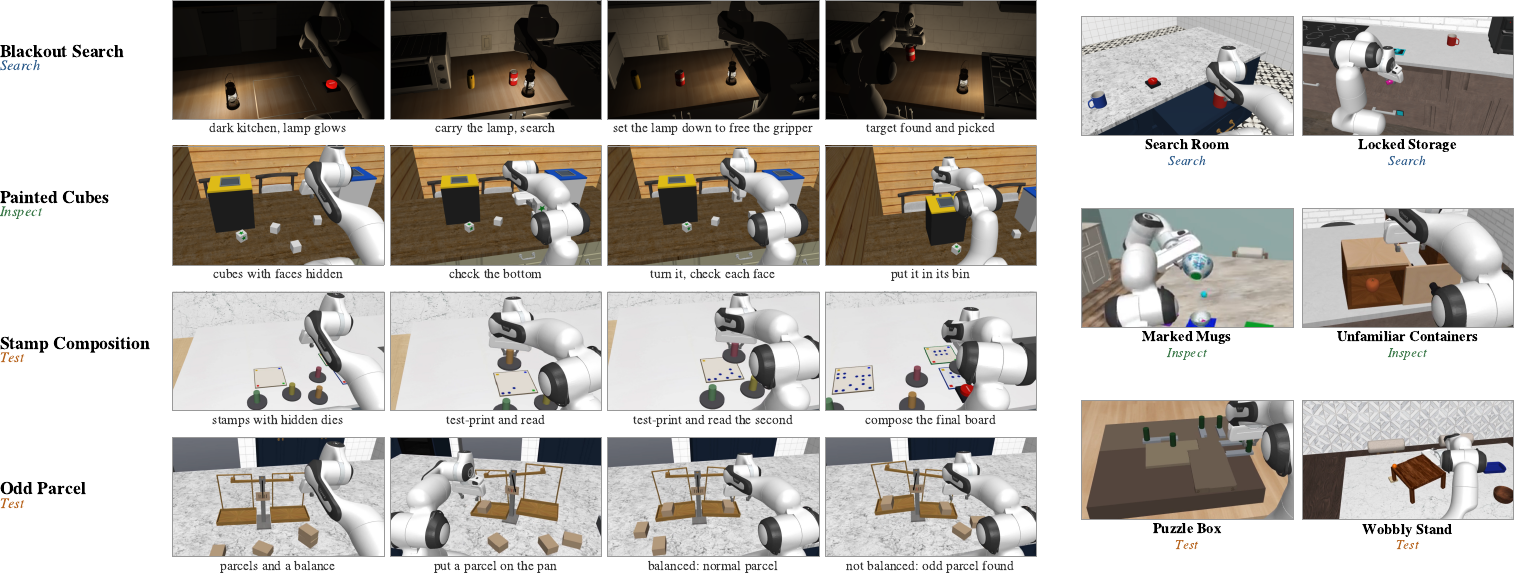}
\caption{\textbf{\dataset{} tasks.} Left: four tasks, from the initial scene through revealing the hidden information to completion. Right: the other six tasks. Colours mark families.}
\label{fig:tasks}
\end{figure}

\paragraph{Tasks.} The ten tasks fall into three families by how the missing information is obtained: three Search tasks, three Inspect tasks, and four Test tasks. \Cref{fig:tasks} shows one demonstration of each task, while \cref{tab:task_overview} lists what each task hides, and \autoref{sec:tasks} describes every task in full, including the parameters its instances vary, with a frame sequence for each in \cref{fig:task_wall}.

\paragraph{Environment.} All tasks run in RoboCasa365 kitchens~\cite{robocasa365} simulated in MuJoCo~\cite{todorov2012mujoco}, with a Franka Panda arm on a mobile base controlled at 20\,Hz and observed by two scene cameras and a wrist camera. Each task places its objects, containers, and mechanisms into a kitchen and states the goal as a natural-language instruction. These are spread over the kitchen, so the robot must drive between them, most of all in the Search tasks. The instruction never gives the hidden information of \cref{tab:task_overview}, which is kept in a private specification inaccessible to the agent.

\begin{table}[t]
\centering
\footnotesize
\caption{\textbf{Tasks at a glance.} What each task hides and how the agent can reveal it. \Cref{tab:capability_matrix} (\autoref{app:capa}) lists the capabilities each task targets.}
\label{tab:task_overview}
\begin{tabular*}{\textwidth}{@{\extracolsep{\fill}}llll@{}}
\toprule
Family & Task & Hidden information & How it is revealed \\
\midrule
\multirow{3}{*}{Search}
 & Locked Storage & target behind locks & find the tokens that unlock each compartment \\
 & Search Room & where the targets are & open storage places until the set is complete \\
 & Blackout Search & targets in the dark & carry a light and search what it illuminates \\
\midrule
\multirow{3}{*}{Inspect}
 & Painted Cubes & marks on unseen faces & turn each cube to see every face \\
 & Marked Mugs & labels under vessels & lift or tilt each vessel to read its underside \\
 & Unfamiliar Containers & how containers open & try each opening mechanism \\
\midrule
\multirow{4}{*}{Test}
 & Puzzle Box & bolt blocking order & try moves and observe which parts move \\
 & Stamp Composition & patterns and rotations & print trial impressions on a test surface \\
 & Wobbly Stand & short legs and gaps & place a ball and watch where it rolls \\
 & Odd Parcel & which parcel is odd & compare groups of parcels on a balance \\
\bottomrule
\end{tabular*}
\end{table}

\paragraph{Episodes and submission.} An episode ends when the robot physically presses a \texttt{SUBMIT} button in the scene. This press freezes the score, so only the state at that moment counts, and the agent itself must judge when its evidence is sufficient. Some actions cannot be undone before that point: a cube dropped into a bin stays there, a token shut inside a compartment is locked away, and ink on the final surface is permanent. An episode that ends without a press counts as a failure.

\paragraph{Instances.} Each instance is generated from a seed and a setting of the task's main parameters. The seed fixes the kitchen, the object layout, the materials, and the remaining details. Five tasks also vary occlusion, with every task object in view at the start (\emph{Visible}), some object outside every camera view (\emph{Look}), or some object under a cover (\emph{Uncover}), as detailed in \autoref{sec:factor_details}. We evaluate on 50 instances per task, which cover every level of the task's main parameters and together span 18 kitchen layouts and 5 kitchen styles. All agents play the same instances.

\paragraph{Scoring.} An episode succeeds if and only if every goal condition holds at the press. For example, every target on the tray or every cube in its correct bin. Since a single wrong unit fails the whole episode, we also measure progress, a per-task partial credit between 0 and 1 defined in \autoref{app:progress}.

\subsection{Demonstration Dataset}
\label{sec:demonstrations}

To support the training of learned policies, we also release a dataset of successful demonstrations for all ten tasks. The demonstrations are generated by scripted oracles, one per task. An oracle reads the private specification to execute reliably, but its exploration is scripted as that of an agent without this knowledge, since a demonstration that goes straight to the answer would teach a policy nothing about exploration. The oracles open compartments nearest-first until the targets are seen, lift and tilt each vessel to read its label, turn each cube face by face, print test impressions before stamping the final board, weigh parcels on the balance, and test with the ball again after placing a shim (\cref{tab:dataset}). In Unfamiliar Containers the oracle visits the boxes nearest-first by route length, and on each box tries the actions its visible knob suggests in a fixed order (a knob on the lid: lift, twist, slide; a knob on a face: pull, press, slide, turn), including on decoy knobs that open nothing; a box that resists twelve tries is left for later. Most episodes therefore contain several failed tries before a box opens.

\paragraph{Size and selection.} The dataset contains 500 episodes for each task, 5{,}000 in total, amounting to 366 hours of interaction at 20\,Hz (\cref{tab:dataset}). Episodes last 4.4 minutes on average, about 5{,}300 control steps, from 1.6 minutes in Puzzle Box to 7.5 minutes in Blackout Search, where the oracle carries a lamp from place to place and searches each one in the dark. The 500 episodes of each task cover its workload levels (\cref{sec:factor_details}) in equal numbers and spread over the kitchen layouts and styles. No demonstration shares an instance with the evaluation set. Moreover, the evaluation holds out the kitchen styles, which set the textures and materials of the cabinets, counters, and walls, as well as some task configurations, such as certain target patterns in Stamp Composition and bolt-chain configurations in Puzzle Box. Thus, a policy trained on these demonstrations is always evaluated on unseen scenes and configurations.

Every episode is verified, and every frame carries a two-level language annotation, a stage and the subtask within it, written from the robot's viewpoint. \autoref{sec:demonstration_details} describes the verification, the annotations, and the release format.

\subsection{Capabilities needed for \dataset{} Tasks}
\label{sec:capabilities}

Completing a \dataset{} task requires two kinds of actions. An information-seeking action reveals information that informs later actions, such as those in the last column of \cref{tab:task_overview}. A task action directly drives physical progress toward the goal. The two roles can overlap. For example, opening a container may reveal its mechanism and make retrieval possible. These actions draw on four groups of capabilities: \emph{evidence acquisition} (finding and revealing the missing information), \emph{evidence use} (integrating and remembering what has been observed), \emph{interactive inference} (inferring hidden structure from the outcomes of interactions), and \emph{action organization} (planning action sequences and anticipating their consequences). \autoref{app:capa} details nine finer capabilities within these groups, and \cref{tab:capability_matrix} marks which of them each task targets.

Our evaluation separates execution from the rest. \Cref{sec:control} tests the execution skills in isolation, and \cref{sec:analysis} attributes each failure to one of four families that map onto these groups. Missing evidence mainly reflects weak evidence acquisition, wrong decisions weak evidence use or interactive inference, and side effects weak action organization, while execution failures reflect control, the part that \cref{sec:control} measures.

\section{Experiments}
\label{sec:experiments}

We evaluate \dataset{} in three parts. First, six frontier multimodal models act as zero-shot robot policies through a common visuomotor interface (\cref{sec:frontier_results}); this is the primary focus of the work. Second, to separate exploration from execution, three of these models are tested on the execution skills the tasks are built from, each in isolation and with the information the full task would have to discover supplied up front (\cref{sec:control}). Third, a $\pi_{0.5}$ policy fine-tuned on our demonstrations serves as a learned-policy baseline. As a memoryless VLA model, it shows how effectively the standard policy-learning approach performs on tasks requiring exploratory interaction (\cref{sec:pi05_results}).

\subsection{Experimental Settings}

\paragraph{Frontier Agents.} Owing to the impressive multimodal capabilities of the current frontier models and their success in robotic control~\cite{zhi2025come,anthropic2026clauderobotics,chen2026astra}, we evaluate six of them zero-shot: \argonfull{}, \gptsfull{}, \opusfull{} (\opus{} below), \solfull{}, and \fablefull{} (\fable{}) with medium thinking, and \geminifull{} with high thinking. Every model plays the same 50 evaluation instances of each task (\cref{sec:setup}), 500 episodes per model.

\paragraph{Interface.} All models run in the same harness, built on the Inspect Robots agent framework~\cite{inspect-robots} and adapted for \dataset{} (\autoref{sec:harness}). At each decision the model receives its proprioception, the elapsed simulated time, the number of decisions remaining, and three $512\times512$ images from two scene cameras and a wrist camera. It replies with exactly one of four tool calls (\cref{tab:tools}): \texttt{ARM} moves the gripper, \texttt{BASE} drives the mobile base, \texttt{WAIT} holds both, and \texttt{STOP} abandons the episode. The model chooses how long each command runs, up to 10 seconds of simulated time, and there are no tools that locate or grasp objects for it. Besides a \texttt{SUBMIT} press, an episode ends when the model stops or replies without a tool call, or after 200 decisions. The conversation accumulates for the whole episode, including the model's earlier reasoning, and only the images are pruned, to the two most recent observations. The system prompt and tools are given in \autoref{sec:prompt}.

\paragraph{VLA Model Evaluation.} To see how far a learned policy gets on these tasks, and because frontier models are often impractical to deploy for reasons of cost and latency, we also evaluate a supervised fine-tuned (SFT) $\pi_{0.5}$ model~\citep{intelligence2025pi}, trained on the demonstrations of \cref{sec:demonstrations} and evaluated on the held-out evaluation instances of \dataset{}. The model takes three camera views (left and right base cameras, plus the wrist camera), 16-D robot proprioception, and the task prompt, predicting a subtask label and 20-step action chunks ($1.0$\,s at 20\,Hz). Training minimizes a combined objective of flow-matching action loss and causal subtask prediction. Unlike the language-model agents, it acts at the control rate, sending arm, gripper, and base commands at 20\,Hz, on the same physics and with the same scoring. At test time, the policy executes closed-loop replanning in the simulator until physical \texttt{SUBMIT} or task timeout. Training and inference procedures are detailed in \cref{sec:vla_details}.

\paragraph{Metrics.} \emph{Success rate} (SR, in \%) is the share of episodes whose goal state held at a valid \texttt{SUBMIT} press. \emph{Progress} (Prog, in \%) is the mean of a per-task, staged measure of how much of the goal state was reached, defined in \autoref{app:progress}; it is computed on the final state of every episode, however it ended, and credits partial results in failed episodes. We also record how each episode ended (correct or wrong submission, stop, text answer, or out of decisions) and, for the frontier models, the tokens, monetary cost at list prices, and wall-clock and simulated time per episode. The isolated skill tests report the success rate of each skill.

\subsection{Frontier Models on \dataset{}}
\label{sec:frontier_results}

\begin{table}[t]
\centering
\small
\caption{\textbf{Main results.} SR: success rate. Prog: mean progress. Both in \%, 50 episodes per cell. Best SR per row in bold; shading darkens from 0 to 100\%. Models are ordered by overall SR. Overall w/o Puzzle Box excludes the easiest task, an outlier on which every model scores far above its other tasks.}
\label{tab:main}
\setlength{\tabcolsep}{5pt}
\resizebox{\linewidth}{!}{%
\begin{tabular}{llrrrrrrrrrrrr}
\toprule
& & \multicolumn{2}{c}{\makecell{Gemini\\4 Argon}} & \multicolumn{2}{c}{\makecell{GPT-6\\Astra}} & \multicolumn{2}{c}{\makecell{Claude\\Opus 5.5}} & \multicolumn{2}{c}{\makecell{GPT-6.1\\Sol}} & \multicolumn{2}{c}{\makecell{Claude\\Fable 5.1}} & \multicolumn{2}{c}{\makecell{Gemini\\3.8 Flash}} \\
\cmidrule(lr){3-4}\cmidrule(lr){5-6}\cmidrule(lr){7-8}\cmidrule(lr){9-10}\cmidrule(lr){11-12}\cmidrule(lr){13-14}
Family & Task & SR & Prog & SR & Prog & SR & Prog & SR & Prog & SR & Prog & SR & Prog \\
\midrule
\multirow{3}{*}{Search}
 & Locked Storage        & \mhb{28.0} & \mh{64.4} & \mhb{28.0} & \mh{52.9} & \mh{6.0}           & \mh{35.2} & \mh{12.0} & \mh{38.5} & \mh{8.0}  & \mh{21.7} & \mh{0.0}          & \mh{9.2}  \\
 & Search Room           & \mhb{6.0}  & \mh{41.9} & \mh{0.0}           & \mh{31.0} & \mh{0.0}           & \mh{26.7} & \mh{0.0}  & \mh{19.4} & \mh{0.0}  & \mh{19.2} & \mh{2.0}          & \mh{17.5} \\
 & Blackout Search       & \mh{0.0}   & \mh{35.9} & \mh{0.0}           & \mh{23.7} & \mh{0.0}           & \mh{17.4} & \mh{0.0}  & \mh{18.5} & \mh{0.0}  & \mh{17.1} & \mh{0.0}          & \mh{4.4}  \\
\midrule
\multirow{3}{*}{Inspect}
 & Painted Cubes         & \mhb{36.0} & \mh{74.6} & \mh{30.0}          & \mh{74.8} & \mh{14.0}          & \mh{59.2} & \mh{18.0} & \mh{69.2} & \mh{14.0} & \mh{61.5} & \mh{2.0}          & \mh{11.3} \\
 & Marked Mugs           & \mh{34.0}  & \mh{71.5} & \mhb{38.0} & \mh{69.7} & \mh{16.0}          & \mh{51.4} & \mh{6.0}  & \mh{37.2} & \mh{12.0} & \mh{31.8} & \mh{0.0}          & \mh{0.8}  \\
 & Unfamiliar Containers & \mh{0.0}   & \mh{34.0} & \mhb{6.0}  & \mh{32.2} & \mh{0.0}           & \mh{17.2} & \mh{0.0}  & \mh{23.1} & \mh{0.0}  & \mh{17.0} & \mh{0.0}          & \mh{6.3}  \\
\midrule
\multirow{4}{*}{Test}
 & Puzzle Box            & \mh{90.0}  & \mh{94.2} & \mhb{96.0} & \mh{99.6} & \mh{68.0}          & \mh{85.5} & \mh{72.0} & \mh{90.9} & \mh{70.0} & \mh{81.0} & \mh{12.0}         & \mh{33.0} \\
 & Stamp Composition     & \mh{2.0}   & \mh{30.2} & \mhb{14.0} & \mh{55.3} & \mh{2.0}           & \mh{39.0} & \mh{8.0}  & \mh{29.4} & \mh{2.0}  & \mh{36.9} & \mh{0.0}          & \mh{3.4}  \\
 & Wobbly Stand          & \mhb{32.0} & \mh{33.0} & \mh{10.0}          & \mh{18.6} & \mh{24.0}          & \mh{32.2} & \mh{4.0}  & \mh{7.3}  & \mh{6.0}  & \mh{14.4} & \mh{2.0}          & \mh{4.6}  \\
 & Odd Parcel            & \mhb{10.0} & \mh{13.0} & \mhb{10.0} & \mh{22.3} & \mh{8.0}           & \mh{9.7}  & \mh{2.0}  & \mh{6.7}  & \mh{2.0}  & \mh{9.0}  & \mh{2.0}          & \mh{2.0}  \\
\midrule
\multicolumn{2}{l}{Overall}                & \mhb{23.8} & \mh{49.3} & \mh{23.2} & \mh{48.0} & \mh{13.8} & \mh{37.3} & \mh{12.2} & \mh{34.0} & \mh{11.4} & \mh{31.0} & \mh{2.0} & \mh{9.2} \\
\multicolumn{2}{l}{Overall w/o Puzzle Box} & \mhb{16.4} & \mh{44.3} & \mh{15.1} & \mh{42.3} & \mh{7.8}  & \mh{32.0} & \mh{5.6}  & \mh{27.7} & \mh{4.9}  & \mh{25.4} & \mh{0.9} & \mh{6.6} \\
\bottomrule
\end{tabular}
}
\end{table}

\begin{figure}[t]
\centering
\includegraphics[width=\linewidth]{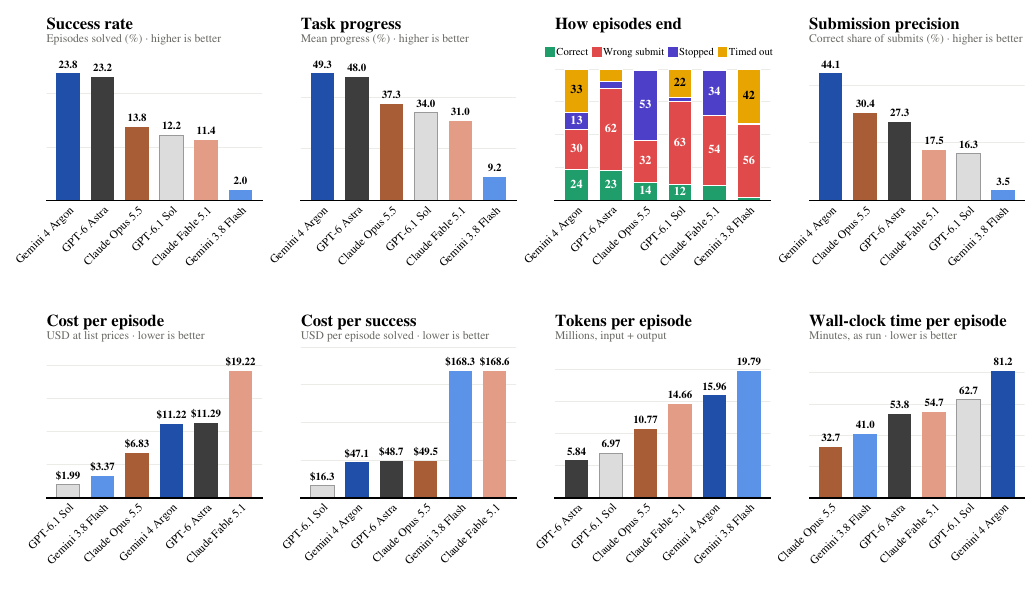}
\caption{\textbf{Results per model at a glance.} Top: success, progress, how the episodes ended, and submission precision, the share of submissions that were correct. Bottom: cost, tokens, and wall-clock time per episode. All values are over the 500 episodes of each model. Bars are ordered from best to worst, except in How episodes end, where ``Stopped'' counts episodes the model ended on its own, by a stop call or a text answer, and ``Timed out'' episodes that ran out of the 200 decisions. Exact values: \cref{tab:main,tab:cost,tab:outcomes}.}
\label{fig:results_overview}
\end{figure}

\Cref{tab:main} and \cref{fig:results_overview} show two models clearly ahead of the rest, \argon{} with a success rate (SR) of 23.8\% and \gpts{} with 23.2\%. \opus{}, \sol{}, and \fable{} follow close to one another (11--14\%), \gemini{} trails at 2.0\%, and mean progress follows the same order. The two open search tasks are hard for every model: no model solves a single Blackout Search episode, and Search Room is solved in only three episodes of \argon{} and one of \gemini{}. On the opposite end, Puzzle Box is the easiest task for every model, 44 to 58 points above the next best task for the five stronger models, so we also report the averages without it. Without Puzzle Box, \argon{} and \gpts{} succeed about 1.9 to 3.3 times as often as \opus{}, \sol{}, and \fable{}. \argon{} leads on Search Room, Painted Cubes and Wobbly Stand, and \gpts{} on four other tasks. We additionally evaluated two open-source models, Qwen3.8-27B and Qwen3.8-Flash (125B parameters, 6B active). However, they failed every full-task episode we ran (details in \autoref{app:open_models}).

\paragraph{Cost Analysis.} Evaluating the six models on 500 episodes each cost about \$27{,}000 in total, from \$997 for \sol{} to \$9{,}610 for \fable{}. \sol{} is by far the cheapest model, both per episode and per successful episode, and \argon{}, \gpts{} and \opus{} cost about three times as much per success (\cref{fig:results_overview}, \cref{tab:cost}). \fable{} and \gemini{} are the least economical per success, \fable{} because it is the most expensive per episode and \gemini{} because it rarely succeeds. The models make a similar number of decisions but spend very different amounts of time, and the two GPT models command the longest robot motions (\cref{tab:cost}).

\paragraph{How Episodes End.} The models give up in different ways (last panel of \cref{fig:results_overview}, \cref{tab:outcomes}). \gpts{} and \sol{} almost always press \texttt{SUBMIT}, but mostly with the task unfinished: only 27\% and 16\% of their submissions are correct. \opus{} and \fable{} instead often stop on their own, and \argon{} and \opus{} are the most precise when they do submit (44\% and 30\%). \gemini{}, \argon{} and \sol{} run out of the 200 decisions far more often than the others. No model stops because it is done, and no stopped episode had its goal state met. Since the remaining decisions are shown at every turn, a stop and a timeout are two reactions to the same state of little budget and little progress. \opus{} stops with a median of 24 decisions left, where \gpts{} submits what it has and \sol{} runs out. We read the two together as giving up.

\subsection{Execution Skills in Isolation}
\label{sec:control}

Task failures on \dataset{} can come from either ineffective exploration or poor manipulation. To separate the two, we test these execution skills in isolation with three of the models, \gpts{}, \opus{}, and \sol{}, and with the two open models Qwen3.8-Flash and Qwen3.8-27B. To isolate execution, we remove the need for exploration by providing information normally uncovered during a task, either in the instruction or by placing a single unambiguous target in the scene. We group the evaluated skills into two categories (\cref{tab:control}). \emph{General skills} are common across the benchmark, including picking and placing objects on a target and driving the base to an item. \emph{Task-specific skills} are operations required by individual tasks, such as opening a drawer or cabinet, opening an unfamiliar container from instructions, turning a cube, and sliding a shim under a stand. Each skill test starts directly in front of the target, omits the \texttt{SUBMIT} requirement, terminates once the goal is reached or after 50 decisions, and uses the same harness as the main evaluation (\autoref{app:control}).

\begin{table}[t]
\centering
\small
\caption{\textbf{Execution skills in isolation.} Success rate (\%), 20 scenes per skill. Best per row in bold. Grey: mean decisions of the successful episodes. Details in \autoref{app:control}.}
\label{tab:control}
\newcommand{\sub}[1]{{\color{black!60}#1}}
\setlength{\tabcolsep}{3pt}
\renewcommand{\arraystretch}{0.9}
\begin{tabular}{@{}lll*{3}{@{\hspace{14pt}}r@{\hspace{3pt}}l}@{\hspace{8pt}}|*{2}{@{\hspace{8pt}}r@{\hspace{3pt}}l}@{}}
\toprule
& Skill & Used in & \multicolumn{2}{c}{\makecell{GPT-6\\Astra}} & \multicolumn{2}{c}{\makecell{Claude\\Opus 5.5}} & \multicolumn{2}{c}{\makecell{GPT-6.1\\Sol}} & \multicolumn{2}{c}{\makecell{Qwen3.8\\Flash}} & \multicolumn{2}{c}{\makecell{Qwen3.8\\27B}} \\
\midrule
\multirow{2}{*}{General}
 & Pick and place (on counter) & All tasks  & \textbf{95} & \sub{(19)} & 90 & \sub{(21)} & \textbf{95} & \sub{(18)} & 15 & \sub{(32)} & 0 & \sub{(--)} \\
 & Drive and pick            & Most tasks & \textbf{100} & \sub{(23)} & \textbf{100} & \sub{(23)} & 90 & \sub{(23)} & 5 & \sub{(27)} & 0 & \sub{(--)} \\
\midrule
\multirow{6}{*}{\makecell[l]{Task-\\specific}}
 & Open drawer/cabinet       & Search tasks & 50 & \sub{(22)} & \textbf{60} & \sub{(31)} & 30 & \sub{(36)} & 0 & \sub{(--)} & 0 & \sub{(--)} \\
 & Pick from drawer/cabinet  & Search tasks & \textbf{70} & \sub{(15)} & 60 & \sub{(17)} & 60 & \sub{(20)} & 30 & \sub{(23)} & 0 & \sub{(--)} \\
 & Open named container & Containers & \textbf{100} & \sub{(20)} & 90 & \sub{(16)} & 95 & \sub{(21)} & 15 & \sub{(35)} & 5 & \sub{(47)} \\
 & Pick from open container  & Containers & \textbf{90} & \sub{(20)} & \textbf{90} & \sub{(24)} & 70 & \sub{(19)} & 5 & \sub{(24)} & 0 & \sub{(--)} \\
 & Flip a cube               & Painted Cubes & 100 & \sub{(12)} & 100 & \sub{(19)} & 100 & \sub{(11)} & 70 & \sub{(34)} & 15 & \sub{(10)} \\
 & Shim the short leg        & Wobbly Stand  & 35 & \sub{(33)} & \textbf{60} & \sub{(32)} & 40 & \sub{(40)} & 0 & \sub{(--)} & 0 & \sub{(--)} \\
\midrule
\multicolumn{3}{@{}l}{All skills} & 80 & \sub{(19)} & \textbf{81} & \sub{(22)} & 72 & \sub{(21)} & 18 & \sub{(31)} & 2 & \sub{(20)} \\
\bottomrule
\end{tabular}

\end{table}

As shown in \cref{tab:control}, all three models demonstrate strong proficiency on general skills. \gpts{} and \opus{} drive to an object and pick it up in every episode and \sol{} in 90\%, while all three place an object on a counter target in 90--95\% of the trials. Performance on task-specific skills is more varied. Certain operations achieve consistently high success, such as turning a cube (100\%) and opening an unfamiliar container when instructed (90--100\%). In contrast, manipulation involving furniture or tight clearances is much harder. Models open a drawer or cabinet in only 30--60\% of episodes, retrieve an object from one in 60--70\%, and slide the shim under the named table leg in 35--60\%. These failures are rarely due to an inability to reach the target. The robot touches the handle in 75--85\% of the drawer and cabinet episodes and brings the shim within 3\,cm of the leg in 90--100\% of trials, demonstrating that the difficulty lies in the final, precise manipulation. Surprisingly, we found that the same models do better on two of these skills inside the full tasks. For example, \gpts{} opens 90\% of the compartments it tries there, against 50\% in isolation, and needs 10 decisions on average instead of 22 (second finding in \cref{par:execution}).

Across all evaluated skills, the three frontier models perform closely (80\%, 81\%, and 72\%). Notably, \opus{} slightly exceeds \gpts{} in isolation, despite \gpts{} succeeding 1.7 times as often on the full benchmark tasks. Performance differences among the frontier models on \dataset{} are therefore not explained by execution alone. The open models fall far behind, completing 18\% (Qwen3.8-Flash) and 2\% (Qwen3.8-27B) of the tests, mostly by turning a cube. Their failure on the full tasks (\autoref{app:open_models}) is therefore mostly due to poor control.

\subsection{Fine-Tuned $\pi_{0.5}$}
\label{sec:pi05_results}
\begin{wrapfigure}{r}{0.45\textwidth}
\vspace{-12pt}
\centering
\includegraphics[width=\linewidth]{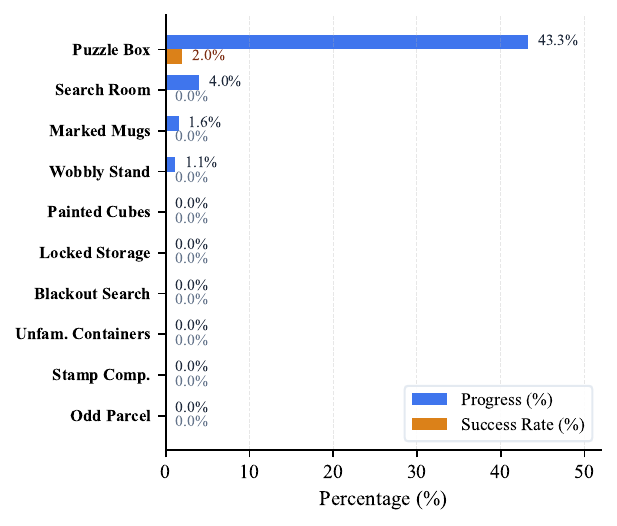}
\vspace{-14pt}
\caption{Fine-tuned $\pi_{0.5}$ evaluation results.}
\label{fig:pi05_results}
\vspace{-8pt}
\end{wrapfigure}
\cref{fig:pi05_results} shows the performance of the fine-tuned $\pi_{0.5}$ models on the same 50 held-out evaluation instances per task as the frontier agents. Overall, $\pi_{0.5}$ performs substantially worse compared to frontier multimodal agents. On Puzzle Box, the policy achieves a 2.0\% success rate (1 of 50 episodes) with an average task progress of 43.3\%, demonstrating that it can occasionally unbolt early sliders but fails to navigate the complete multi-stage dependency to extract the object. Across the remaining tasks, performance collapses entirely, recording 0.0\% success and near-zero progress, with most episodes failing due to timeout.

This failure highlights the challenge of out-of-distribution (OOD) generalisation in goal-directed exploration. The evaluation instances used novel kitchen styles, materials, task mechanisms and target configurations that are not present in the training dataset (\cref{sec:demonstrations}). Thus, the benchmark cannot be solved by simply memorizing state-action mappings or trajectory sequences from training data. Because standard VLA architectures such as $\pi_{0.5}$ lack explicit episodic memory, minor execution errors in OOD scenes quickly compound, leaving the policy stuck in repetitive loops or idling until the decision horizon expires.

\section{Failure Analysis}
\label{sec:analysis}

\Cref{sec:experiments} showed that frontier models fail in most benchmark episodes, even though they reliably execute the underlying physical skills when hidden information is supplied (\cref{sec:control}). However, isolated skill tests cannot determine what fraction of failures are caused by execution errors, nor identify what causes the remaining failures. In this section, we attribute each failure of \gpts{}, \opus{}, and \sol{} to its initial point of failure: missing evidence, a wrong decision, a disturbed scene, or a failed action. Interpreting these results alongside the isolated skill tests reveals that the two analyses are consistent: most failures arise from exploration and decision steps, not from execution of skills.

\subsection{Attributing Failures}
\label{sec:attributing}

To complete a task, an agent repeatedly (i) acquires the evidence it needs, (ii) interprets it and decides what to do (including when to stop), and (iii) executes the decision, observing outcomes and repairing mistakes. We classify failures by the earliest unrecovered breakdown in this loop into four families, two of which arise during physical action:
\begin{itemize}
    \item \textbf{Missing evidence}: the evidence required to fulfill a task goal never reached the agent, not because a physical attempt failed, but because it was never explored. For example, the agent stops or commits to a decision with a hiding place never opened, a cube face never turned to a camera, or a parcel never weighed.
    \item \textbf{Wrong decision}: the necessary evidence appeared in the agent's observations, but the agent either acted against it (e.g., misreading a label, or stamping the wrong cell after a test print) or failed to act on it (e.g., spotting a target but never retrieving it, or stopping while unfinished work remained in observation).
    \item \textbf{Side effect}: the action itself succeeded but disturbed a task-critical object or state without recovery, e.g., tipping a mug to read its label and spilling balls onto the floor, or accidentally knocking over a stand. Whether the agent failed to foresee the consequence or accidentally collided with the object cannot be determined from the data.
    \item \textbf{Execution failure}: the decided action was appropriate, but its physical execution failed, e.g., slipping off a handle when trying to open a compartment, missing a grasp, dropping an object, placing an item off-target, or failing to recover a dropped item.
\end{itemize}
\autoref{sec:attribution} gives the rules at the boundaries between the families.

\begin{figure}[t]
\centering
\includegraphics[width=\linewidth]{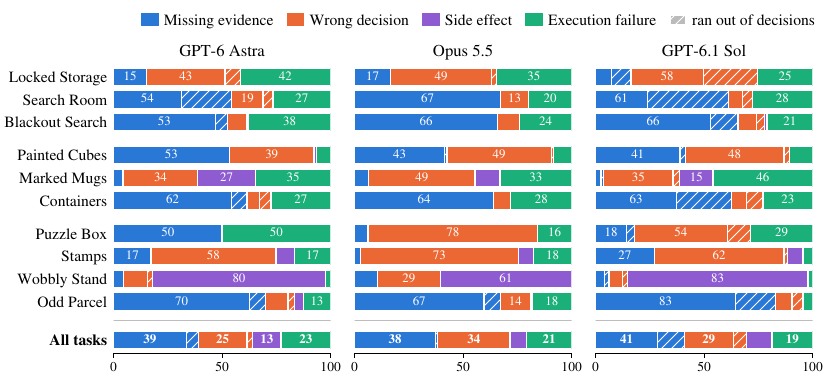}
\caption{\textbf{Failure attribution by family,} in \% of failures, with each failed episode weighted equally, per task and over all tasks. Containers: Unfamiliar Containers. Stamps: Stamp Composition. Counts in \cref{tab:failure_per_task}.}
\label{fig:failure_breakdown}
\end{figure}

\paragraph{Units and procedure.} One episode can fail in several ways at once, so we label every \emph{unmet requirement} of a failed episode as a separate unit, not just the first. If a Search Room episode leaves two of three targets off the tray, one never found and one dropped on the way, it gives two units: the first is labelled missing evidence, the second an execution failure. Each unit is labelled at the first unrepaired error on its own path, the sequence of events that led to that requirement's final state. Each task defines an ordered list of deterministic rules evaluated on every unit. Each rule tests a verifiable condition, such as whether a compartment was opened, whether an object was visible in the camera observations, or whether a released object landed on the floor. The first rule that matches assigns the failure family to that unit. \autoref{sec:attribution} gives the shared measurements and thresholds. The rules of every task and the label of every unit are released with the benchmark. We consider evidence to be observed if it is visible in at least one camera view received by the model. Each failed episode carries equal weight, split over its units, so the episode above counts one half towards each of the two families.

\Cref{fig:failure_breakdown} shows the result for 931 / 1143 / 1182 units from 384 / 431 / 439 failed episodes of \gpts{} / \opus{} / \sol{}; values for the three models are given in this order throughout. Missing evidence is the largest family for every model (39 / 38 / 41\%), followed by wrong decisions (25 / 34 / 29\%); together they account for 64--72\% of the failures. Execution failures are 23 / 21 / 19\% and side effects 13 / 7 / 12\%. The hatched parts are failures in episodes that ran out of decisions (8 / 1 / 19\%). For \gpts{} and \sol{}, about half of these are hiding places that were never opened. \Cref{sec:findings} reads the evidence behind these shares as four failure patterns.

\subsection{Failure Patterns}
\label{sec:findings}

\paragraph{Incomplete exploration is the main source of failure.} Most often, the models stop exploring while parts of the scene remain uninspected. In about half of the failures (43 / 51 / 49\%), an object the task needed was never placed (\cref{tab:end_commit}). In most of these cases (23--25\% of all failures), the object remained hidden inside a drawer, cabinet, box, or cover that was never opened. Crucially, the hiding place itself was rarely missed. It typically appeared in the agent's observations five or more times, while hiding places that were never observed account for only 1--3\% of failures. In 15--22\% of the failures, the target, or the token that unlocks it, was visible and observed by the agent but never acted on: e.g., a target never fetched, or a dropped object or spilled ball left on the counter.

Beyond stopping exploration early, models also commit to decisions prematurely without gathering sufficient evidence. Roughly a tenth of all failures (9--11\%) occurred when models committed objects to locations before observing what was needed to decide correctly. In Odd Parcel, most wrongly selected parcels (71--84\%) were boxed before weighings could reveal whether they were odd. In Painted Cubes, marks on uninspected cube faces remained unseen, causing models to misclassify qualifying cubes far more often than unqualified ones (22--37\% against 5--10\%; \cref{fig:cubes_direction}). In Stamp Composition, 20--23\% of incorrectly placed stamps by \gpts{} and \sol{} were printed without a prior test print (compared to 3\% for \opus{}).

Controlled experiments varying visual occlusion corroborate this exploration bottleneck (\cref{tab:factors}). When task objects start out of view or under a cover, overall success drops by 8--10 points. Crucially, these additional failures fall entirely under \emph{missing evidence}, while wrong decisions, execution failures, and side effects do not increase (\cref{tab:factor_phases}). In other words, hiding an object makes agents less likely to find it, but no worse at manipulating it once found.

\begin{table}[h!]
\centering
\footnotesize
\hfill\begin{minipage}[t]{0.37\linewidth}\centering
\setlength{\tabcolsep}{4pt}
\begin{tabular*}{\linewidth}{@{\extracolsep{\fill}}lrrr@{}}
\toprule
& \makecell{GPT-6\\Astra} & \makecell{Opus\\5.5} & \makecell{GPT-6.1\\Sol} \\
\midrule
\multicolumn{4}{@{}l}{\emph{Open drawer/cabinet}} \\
\quad Isolated & 50 & 60 & 30 \\
\quad Full task & 90 & 78 & 75 \\
\quad $\Delta$ & $+$40 & $+$18 & $+$45 \\
\addlinespace[1pt]
\multicolumn{4}{@{}l}{\emph{Take target out}} \\
\quad Isolated & 70 & 60 & 60 \\
\quad Full task & 79 & 94 & 81 \\
\quad $\Delta$ & $+$9 & $+$34 & $+$21 \\
\addlinespace[1pt]
\multicolumn{4}{@{}l}{\emph{Shim under a leg}} \\
\quad Isolated & 35 & 60 & 40 \\
\quad Full task & 58 & 60 & 32 \\
\quad $\Delta$ & $+$23 & 0 & $-$8 \\
\bottomrule
\end{tabular*}
\end{minipage}\hfill
\begin{minipage}[t]{0.37\linewidth}\centering
\setlength{\tabcolsep}{4pt}
\begin{tabular*}{\linewidth}{@{\extracolsep{\fill}}lrrr@{}}
\toprule
& \makecell{SR\\(\%)} & \makecell{Prog\\(\%)} & \makecell{Decisions\\(mean)} \\
\midrule
\multicolumn{4}{@{}l}{\emph{\gpts{}}} \\
\quad No cover & 94.4 & 99.6 & 84.7 \\
\quad Covered & 96.4 & 99.5 & 119.5 \\
\quad $\Delta$ & $+$2.0 & $-$0.1 & $+$34.8 \\
\addlinespace[1pt]
\multicolumn{4}{@{}l}{\emph{\opus{}}} \\
\quad No cover & 94.4 & 99.7 & 94.4 \\
\quad Covered & 36.8 & 67.4 & 175.7 \\
\quad $\Delta$ & $-$57.7 & $-$32.3 & $+$81.2 \\
\addlinespace[1pt]
\multicolumn{4}{@{}l}{\emph{\sol{}}} \\
\quad No cover & 72.2 & 90.7 & 105.8 \\
\quad Covered & 60.1 & 87.8 & 150.3 \\
\quad $\Delta$ & $-$12.1 & $-$2.9 & $+$44.5 \\
\bottomrule
\end{tabular*}
\end{minipage}\hfill\null
\caption{\textbf{Left: skills in isolation and in the full tasks.} Success (\%) in the isolated tests (\cref{sec:control}) and in the full tasks (\cref{tab:skills_in_context}). In the full tasks, taking the target out counts only targets the model tried to grasp. \textbf{Right: Puzzle Box with the locks hidden.} Success rate, progress, and mean decisions per episode, on chain lengths 4 and 5.}
\label{tab:skills_puzzle}
\end{table}

\paragraph{Execution is imperfect but not the bottleneck.}\label{par:execution} Execution failures account for 19--23\% of all errors, led by off-target placements and drops to the floor. Surprisingly, opening drawers and cabinets, the skill that fails most in isolation, works better in the full tasks. We analysed the compartments holding a target in Search Room and Locked Storage, and find that models open 75--90\% of those they attempt, compared to 30--60\% in the isolated test (\cref{tab:skills_puzzle}, left). At the same time, the successful opening takes fewer decisions, 10 / 26 / 15 on average, against 22 / 31 / 36 in the isolated test (\cref{tab:skills_in_context}). Possibly, the robot has already adapted to the controls in the earlier steps of a full episode, while each isolated test starts from scratch. The other two skills that fail most in isolation improve or hold up in the full tasks as well. When trying to grasp a target in an opened compartment, the models take it out in 79--94\% of cases, above the isolated 60--70\%. In Wobbly Stand, models place a shim under the short leg in 32--60\% of the full episodes, where they must identify that leg themselves, against 35--60\% in isolation, where the leg is named.

\paragraph{Scene disturbances are rarely prevented or repaired.} Side effects account for 7--13\% of all errors, representing failures that isolated skill tests cannot capture. In these cases, the model's intended action succeeds, but inadvertently disrupts a task-critical state. Roughly two-thirds of these errors involve objects falling to the floor where recovery is impossible, such as the test ball in Wobbly Stand rolling off the stand, or balls spilling when a mug in Marked Mugs is tilted to read its label. Emptying the balls is a viable strategy for inspecting the label, and all three models employ it, but they rarely take preventative measures. An empty bowl was provided in each scene for safely holding the balls, but it was used in only 2--8 of the 50 Marked Mugs episodes and never in Wobbly Stand. Furthermore, only a third of the balls that left their mug or the stand were back in place at the end: 30\% fell to the floor, 24\% stayed on the counter and were never retried, and 13\% were retried without success.

\paragraph{Interactive trial and error remains challenging.} In Puzzle Box, a chain of sliding bolts locks the lid. Without the cover, the blocking dependencies between bolts are directly visible. On the other hand, with the cover, the unlocking sequence must be learned by trying bolts and observing what moves. On matching chain lengths (\cref{tab:skills_puzzle}, right), adding the cover barely affects \gpts{}, which succeeds in 94.4\% of episodes without it and 96.4\% with it (using 120 vs.\ 85 decisions). In contrast, \opus{} drops sharply from 94.4\% to 36.8\%, while \sol{} falls moderately from 72.2\% to 60.1\%. \fable{}, tested on the same instances, drops as much as \opus{}, from 100\% to 34.6\%, using 175 instead of 107 decisions. Roughly half of \opus{}'s failed attempts involve pulling a blocked bolt three or more times without updating its plan (12 of 26). The Unfamiliar Containers task requires a similar capability. Given instructions on the opening mechanism, \opus{} succeeds in 90\% of isolated trials (\cref{tab:control}), but when left to discover the mechanism interactively, it opens only 45\% of attempted containers in the full task, against 70\% and 78\% for \gpts{} and \sol{}.

\FloatBarrier
\section{Conclusion}

The importance of \dataset{} is highlighted by the empirical performance and cost of six frontier models. All six generalist frontier physical agents succeed on only a small fraction of episodes: \argon{} and \gpts{} achieve the highest success rates at 23.8\% and 23.2\%, \opus{}, \sol{}, and \fable{} follow at 11 to 14\%, and \gemini{} achieves the lowest at 2\%; \sol{} is the most economical per successful episode. Two smaller open models fail every full-task episode we ran. Beyond aggregate performance, two complementary analyses show where goal-directed exploration breaks down. When tested in isolation, models succeed on execution skills in 72--81\% of attempts, while our failure analysis attributes only 19--23\% of failures in the full tasks to execution. The two analyses agree that the bottleneck lies in exploring and decision-making rather than in execution, with most failures resulting from exploration that stops too early. The models stop exploration while parts of the scene remain uninspected, or commit to decisions before acquiring necessary evidence (9--11\% of failures). Our controlled occlusion experiments confirm that hiding objects increases missing-evidence errors. Furthermore, physical exploration can have irreversible consequences, disrupting task-critical state in 7--13\% of failures, while interactive hypothesis testing remains brittle: covering the locks in Puzzle Box leaves \gpts{} unaffected but cuts \opus{}'s success from 94\% to 37\%. Finally, models diverge in how they end tasks, either submitting prematurely or exhausting their decision budget. Together, these results demonstrate that progress on embodied exploration requires not only stronger perception and control, but also systematic evidence acquisition, persistent goal and evidence tracking, causal hypothesis testing, and robust monitoring and recovery during physical interaction.

\subsection*{AI use statement}

In this work, we used generative AI tools to provide feedback on research methodology and experiments, assist with method implementation, create or modify scientific figures, summarise existing literature, and draft or edit portions of the manuscript for clarity. We did not use these tools to develop theoretical models, formulate or prove mathematical claims, propose or refine hypotheses, clean or reformat datasets, generate synthetic data, or translate the manuscript. We manually checked LLM-generated code for correctness, verified literature summaries against the cited sources, and reviewed all AI-generated text and figures. We take responsibility for the final content of this work, including text, claims or artifacts produced with the aid of generative AI.

\bibliography{iclr2027_conference}
\bibliographystyle{iclr2027_conference}

\clearpage
\appendix
\addtocontents{toc}{\protect\setcounter{tocdepth}{1}}
\renewcommand{\contentsname}{Appendix Contents}
{\small\tableofcontents}
\bigskip

\section{System Prompt and Tools}
\label{sec:prompt}

Every model receives the same system prompt at the start of the episode, and only the goal text differs between tasks. The prompt for Puzzle Box reads:

\begin{quote}\small
You control a simulated mobile manipulation robot. Use only the provided camera images, robot proprioception and action/budget feedback. Determine what the goal requires from the scene. Choose exactly one arm, base, wait or stop tool per response. There are no semantic object-location or grasp tools. Images are upright RGB from three fixed cameras; new views require physical robot or object motion. The arm tool controls the right gripper site in world metres, with optional world orientation as quaternion xyzw. World z is up. Gripper -1 opens, +1 closes and 0 keeps its current setpoint. The base tool takes normalized body-frame forward, left and yaw velocity commands; positive yaw turns left. The arm stays relative to the base during base motion. The torso is held. A tick is 0.05 seconds. Commands consume up to 200 ticks each, ending early on the first physical \texttt{SUBMIT} button press. Check observed outcomes: commanded targets are not guaranteed to be reached. Physically press \texttt{SUBMIT} to commit the result; the first press freezes the score and ends the episode immediately. Wrong submission, timeout without submission, or calling stop without submission fails. The stop tool abandons the episode without advancing time. Invalid or multiple tool calls perform no physics and still consume this model response. No intermediate success is awarded. Goal: Open the wooden puzzle box on the counter, take out the tangerine inside, and stand it inside the tray. The lid slides sideways but is locked by sliding bolts with round black handles: a part moves only when nothing blocks it, so work out the order. Any order that works is fine. Once the tangerine is inside the tray and you have let go of everything, press the red \texttt{SUBMIT} button. Your first submission ends the episode; a wrong submission or no submission fails.
\end{quote}

\Cref{tab:tools} lists the four tools provided to the agents. After each tool call, the agent receives a response indicating whether the command was accepted, the current simulation tick count, and any error message. To prevent any information leakage, this response never exposes simulator state, runtime exceptions, or task progress.

\begin{table}[h]
\centering
\small
\caption{\textbf{Action tools.} Positions are in world metres, and a tick is 0.05\,s of simulated time. Every tool except \texttt{STOP} accepts a tick count from 1 to 200.}
\label{tab:tools}
\setlength{\tabcolsep}{5pt}
\begin{tabular}{lp{0.62\linewidth}r}
\toprule
Tool & Arguments & Default ticks \\
\midrule
\texttt{ARM} & gripper-site position (required), orientation as a quaternion, and gripper command in $[-1,1]$, where $-1$ opens and $+1$ closes & 120 \\
\texttt{BASE} & body-frame forward, left, and yaw velocity, each in $[-0.5, 0.5]$ & 20 \\
\texttt{WAIT} & optional gripper command, holds the arm and base & 20 \\
\texttt{STOP} & none, abandons the episode, which fails & -- \\
\bottomrule
\end{tabular}
\end{table}

\section{Evaluation Harness}
\label{sec:harness}

\paragraph{Overview.} The evaluation harness extends the Inspect Robots agent framework~\cite{inspect-robots}, which provides the model client interfaces and the turn-based execution loop. All remaining components, such as the simulation environments and task scoring, are specific to \dataset{}.

\paragraph{Observations.} At each decision step, the model receives observations as a user message. The robot's proprioceptive state and the remaining decision budget are provided as text input, along with three $512\times512$ RGB images captured from the left and right scene cameras and the wrist-mounted camera.

\paragraph{Actions and execution.} At each step, the model must reply with exactly one tool call and specify the action's duration in simulation ticks. An \texttt{ARM} command servos the gripper toward the target pose using the robot's operational-space controller. A \texttt{BASE} command applies body-frame velocities at each tick while holding the arm fixed relative to the base, and \texttt{WAIT} holds both. Each command runs for exactly its requested ticks, with no additional settling time. Commands terminate early only if the physical \texttt{SUBMIT} button is pressed.

\paragraph{Memory.} Each episode begins with a fresh conversation context, with no memory retained across episodes. Within an episode, each turn appends the observation, the model's full response (including its reasoning), and the tool execution result to the dialogue history. While the complete textual history is preserved, the image history is pruned, retaining only the two most recent observations consisting of six images in total.

\paragraph{Budgets and termination.} Each episode allows a budget of 200 model decisions, with individual commands running up to 200 simulation ticks. An episode terminates upon pressing the \texttt{SUBMIT} button, exhausting the decision budget, calling \texttt{STOP}, or returning an invalid response. Tasks are scored strictly on the simulator state at the time of episode terminations.

\begin{table}[h]
\centering
\footnotesize
\caption{\textbf{Cost and effort per episode.} Sim / Wall: simulated / wall-clock minutes. In / Out: tokens in millions. Costs in USD at list prices. Best per column in bold (the lowest value, or the highest cache rate). For Gemini 4 Argon, costs exclude the cache-priming calls that worked around a change in Gemini's implicit caching (\$16.76 per episode with them).}
\label{tab:cost}
\setlength{\tabcolsep}{4pt}
\resizebox{\linewidth}{!}{%
\begin{tabular}{lcccccccc}
\toprule
& \multicolumn{3}{c}{Effort per episode} & \multicolumn{3}{c}{Tokens per episode} & \multicolumn{2}{c}{Cost (USD)} \\
\cmidrule(lr){2-4}\cmidrule(lr){5-7}\cmidrule(lr){8-9}
Model & Decisions & Sim min. & Wall min. & In (M) & Out (M) & Cached (\%) & \$/episode & \$/success \\
\midrule
Gemini 4 Argon   & 141.1 & \textbf{4.3} & 81.2 & 15.84 & 0.12 & 86.5 & 11.22 & 47.1  \\
GPT-6 Astra      & 143.7 & 10.4 & 53.8 & \textbf{5.82} & \textbf{0.02} & 91.8 & 11.29 & 48.7  \\
Claude Opus 5.5  & \textbf{143.0} & 5.4 & \textbf{32.7} & 10.70 & 0.07 & 93.5 & 6.83 & 49.5  \\
GPT-6.1 Sol      & 157.4 & 11.2 & 62.7 & 6.94  & \textbf{0.02} & 92.4 & \textbf{1.99} & \textbf{16.3} \\
Claude Fable 5.1 & 148.8 & 6.1  & 54.7 & 14.55 & 0.11 & \textbf{94.2} & 19.22 & 168.6 \\
Gemini 3.8 Flash & 160.8 & 5.1 & 41.0 & 19.66 & 0.13 & 89.3 & 3.37  & 168.3 \\
\bottomrule
\end{tabular}%
}
\end{table}

\section{$\pi_{0.5}$ Fine-Tuning and Evaluation Details}
\label{sec:vla_details}

We fine-tune $\pi_{0.5}$ on each task separately, using that task's demonstrations from our dataset (\cref{sec:demonstrations}).

\paragraph{Observations.} At each decision step, the model receives three RGB camera views resized to $224\times224$ pixels, the robot's proprioceptive state, and the task instruction.

\paragraph{Action Space and Horizon.} The policy predicts action chunks of horizon $H=20$, corresponding to $1.0$\,second of continuous execution at the simulator's $20$\,Hz control frequency. Each predicted action is a 12-dimensional vector consisting of a 6-D end-effector delta pose, a 1-D continuous gripper action ($\pm 1$), 3-D mobile base velocities, and a 2-D torso pose.

\paragraph{Training Objective.}
Our training simultaneously optimizes an autoregressive language modelling objective for subtask prediction and a conditional flow-matching objective for low-level action chunking:
\begin{equation}
    \mathcal{L}_{\text{total}} = \mathcal{L}_{\text{subtask}} + 10 \cdot \mathcal{L}_{\text{flow}},
\end{equation}
where $\mathcal{L}_{\text{subtask}}$ is the cross-entropy loss over causal subtask tokens, and $\mathcal{L}_{\text{flow}}$ is the flow-matching loss conditioned on vision, state, prompt, and the teacher-forced subtask.

\paragraph{Hyperparameters.}
\Cref{tab:vla_hyperparameters} lists the training and inference hyperparameters used. We perform full-parameter fine-tuning with the AdamW optimizer, cosine learning rate scheduling, and exponential moving average (EMA) weight tracking.

\begin{table}[h]
\centering
\small
\caption{\textbf{$\pi_{0.5}$ Training and Inference Hyperparameters.}}
\label{tab:vla_hyperparameters}
\setlength{\tabcolsep}{8pt}
\begin{tabular}{llr}
\toprule
Category & Hyperparameter & Value \\
\midrule
\multirow{9}{*}{Training}
 & Training Steps (per task) & 10{,}000 \\
 & Optimizer & AdamW ($\beta_1=0.9, \beta_2=0.999$, $\epsilon=10^{-8}$) \\
 & Weight Decay & $1\times 10^{-4}$ \\
 & Peak Learning Rate & $2.5\times 10^{-5}$ \\
 & Learning Rate Schedule & Cosine decay with 1000 warmup steps \\
 & Global Batch Size & 32 \\
 & Gradient Clipping Norm & $1.0$ \\
 & Flow Loss Multiplier & $10.0$ \\
 & Action Normalization & Quantile ($q_{01}, q_{99}$) \\
\midrule
\multirow{3}{*}{Inference}
 & Replanning Frequency & Once per 20 ticks (1.0\,s) \\
 & Episode Limit & 12{,}000 control steps (10\,min) \\
 & Flow Denoising Steps & 10 \\
\bottomrule
\end{tabular}
\end{table}

During evaluation, the policy runs in closed loop, replanning every 20 ticks ($1.0$\,s) until the episode terminates upon pressing the physical \texttt{SUBMIT} button or exceeding the task horizon.

\paragraph{Fit on the training distribution.} We evaluate the fine-tuned policy open-loop on 2{,}048 frames sampled from the first 100 Puzzle Box training episodes. For each frame, the policy predicts an action chunk, which we compare with the demonstration's chunk in raw action units. The policy first predicts the current subtask as text and then the action chunk conditioned on it, so we report both its own predicted subtask and the correct subtask from the demonstration. \Cref{tab:pi05_fit} compares the mean squared error with that of always predicting the mean training action.

\begin{table}[h]
\centering
\small
\caption{\textbf{Open-loop fit of the fine-tuned $\pi_{0.5}$} on 2{,}048 Puzzle Box training frames.}
\label{tab:pi05_fit}
\begin{tabular}{lrr}
\toprule
Prediction & Action-chunk MSE & Pooled $R^2$ \\
\midrule
Mean training action & 0.1249 & -- \\
Policy, own predicted subtask & 0.00436 & 0.965 \\
Policy, correct subtask & 0.00434 & 0.965 \\
\bottomrule
\end{tabular}
\end{table}

\section{Benchmark Tasks}
\label{sec:tasks}

\begin{figure}[t]
\centering
\includegraphics[width=\textwidth]{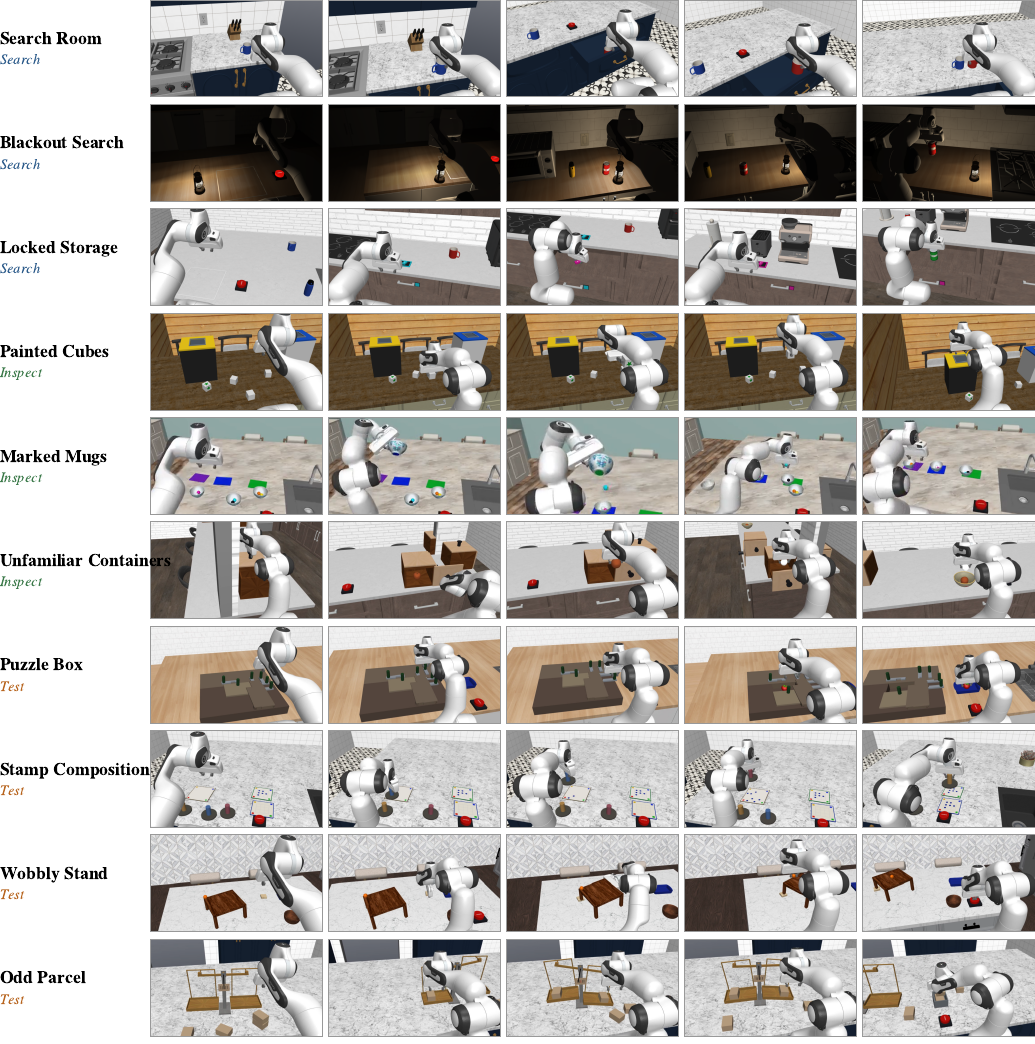}
\caption{\textbf{All ten \dataset{} tasks.} Five frames from one scripted demonstration of each task, from a scene camera, running from the initial scene through the actions that reveal the hidden information to the completed task. Family labels are given in colour.}
\label{fig:task_wall}
\end{figure}

\dataset{} organizes evaluation tasks into three families according to the dominant information-gathering behavior required during execution: \emph{Search \& Explore}, \emph{Object Inspect}, and \emph{Testing}.

\subsection{Search \& Explore}

\paragraph{Search Room.}
A set of target items is stored in open or closed locations around a room,
among distractor items of similar kind. The goal is to collect the complete
set on the tray and press \texttt{SUBMIT}. The locations, the number of storage places,
and which storage places are empty are not given. The robot must choose where
to search, use negative findings to redirect, remember what it has checked,
and stop once the set is complete.
\par\noindent\emph{Main parameters:} 4, 6, or 8 compartments and 2 or 3 targets.
\par\noindent\emph{Goal text:} ``Collect the $\langle$\emph{targets}$\rangle$ on the tray, then press \texttt{Submit}. They may be anywhere in this kitchen; only compartments with handles can be opened. Other objects are not part of the task; you may move them if needed.''

\paragraph{Locked Storage.}
Storage compartments open only when a matching access token rests on their
reader. Tokens are either outside the initial observation or hidden inside
other storage, and some compartments contain tokens for other compartments,
forming a dependency chain. The goal is to retrieve the target item. Finding
a token changes which places can be searched next. Closing a compartment with
a token still inside locks that token away permanently, so the robot must plan
token handling as well as search order.
\par\noindent\emph{Main parameters:} chain depth 1, 2, or 3.
\par\noindent\emph{Goal text:} ``Collect the $\langle$\emph{target}$\rangle$ on the tray, then press \texttt{Submit}. It may be anywhere in this kitchen; some compartments may not open, and a compartment with a coloured lock plate opens only while the token of that colour rests on its reader pad. Other objects on the counter are not part of the task; you may move them if needed.''

\paragraph{Blackout Search.}
Search Room with the lights out. The kitchen stays dark for the whole episode.
A lit lantern stands on a counter and lights about a metre around itself.
Beyond it, the cameras see only the lantern's glowing glass and the
\texttt{SUBMIT} button. The robot searches by carrying the lantern and observes
through all three cameras as in every task. Because the gripper that holds the
lantern cannot also open a drawer, the robot must set the lantern down beside
whatever it works on.
\par\noindent\emph{Main parameters:} 4, 6, or 8 compartments and 2 or 3 targets.
\par\noindent\emph{Goal text:} ``The kitchen is dark. Collect the $\langle$\emph{targets}$\rangle$ on the tray, then press \texttt{Submit}. They may be anywhere in this kitchen; some compartments may not open. A lantern with a glowing glass stands on a counter. Other objects on the counter are not part of the task; you may move them if needed.''

\subsection{Object Inspect}

\paragraph{Marked Mugs.}
Several identical vessels, all mugs or all bowls in a given instance, each hold loose contents, and a label on the
underside of each vessel specifies its destination. The goal is to place
every vessel upright at its destination with its original contents inside.
Reading a label requires lifting or tilting the vessel, which spills the
contents. Spilled contents stay nearby and must be returned to the correct
vessel before submission.
\par\noindent\emph{Main parameters:} 3, 4, or 5 vessels and occlusion Visible, Look, or Uncover.
\par\noindent\emph{Goal text:} ``There are $\langle$\emph{mugs or bowls}$\rangle$ on the counters in this kitchen. Each one has a coloured label on its bottom. Put every one upright on the pad of the same colour, with its own two balls inside, then press \texttt{Submit}. Other objects on the counter are not part of the task; you may move them if needed.''

\paragraph{Painted Cubes.}
A collection of similar objects each carries a hidden property distributed
over its faces, such as the number of painted faces, with some faces turned
toward the table or away from every camera. The goal is to sort every object
by a stated property, for example exactly one painted face, with those that
satisfy it in one bin and the rest in the other.
Partial views can reject an object but cannot confirm it, so the robot must
decide how much inspection each object needs.
\par\noindent\emph{Main parameters:} 4, 6, or 8 cubes and occlusion Visible, Look, or Uncover.
\par\noindent\emph{Goal text:} ``There are $\langle$\emph{cubes}$\rangle$ on the counters in this kitchen. Some of them satisfy this rule: every cube $\langle$\emph{rule}$\rangle$. Put every cube that satisfies it in the blue bin and every other cube in the yellow bin, then press \texttt{Submit}. Other objects on the counter are not part of the task; you may move them if needed.''

\paragraph{Unfamiliar Containers.}
Several identical closed boxes each carry two knobbed panels that look the
same. One panel opens the box by one of eleven mechanisms, such as a hinged
lid, sliding lid, drawer, or latched flip-top. The other is fixed to a solid
wall, and a decoy box has no working panel. Two to four items lie each in its
own box, and all of them must go in the bowl. The robot knows neither which
boxes hold the items nor how each box opens and must discover each mechanism
by interaction.
\par\noindent\emph{Main parameters:} 2, 3, or 4 items and occlusion Visible or Look.
\par\noindent\emph{Goal text:} ``There are $\langle$\emph{boxes}$\rangle$ on the counters in this kitchen. $\langle$\emph{Items}$\rangle$ are inside them. Put $\langle$\emph{all of them}$\rangle$ in the bowl, then press \texttt{Submit}.''

\subsection{Testing}

\paragraph{Stamp Composition.}
Several stamps with unmarked housings each carry a hidden pattern and start at
an unknown rotation. A target pattern is shown on a reference. A test surface
allows trial impressions, and a final surface must end with exactly the target.
The robot must test stamps, read the results, select a subset and their
rotations, compose the target, and press \texttt{SUBMIT}. Marks on the final surface are
permanent.
\par\noindent\emph{Main parameters:} 2, 3, or 4 stamps and occlusion Visible, Look, or Uncover.
\par\noindent\emph{Goal text:} ``There are $\langle$\emph{stamps}$\rangle$ on the counters in this kitchen. Stamp the final board so it shows exactly the pattern on the reference card, then press \texttt{Submit}. The final board is the blue-bordered one with the printed 3x3 grid; the green-bordered reference card is not for stamping. The round self-inking stamps stand face down on the work surface. Each prints a fixed pattern of dots in whatever direction it is held, and the unmarked housings show neither the pattern nor which way it points, so try them on the tan scratch board. Ink stays where it lands, on the scratch board and on the final board alike. Fill exactly the reference card's grid cells with no ink outside the final board's grid, put every stamp back down on the counter and let go, then press the red \texttt{SUBMIT} button. Your first submission ends the episode; a wrong submission or no submission fails. Other objects on the counter are not part of the task; you may move them if needed.''

\paragraph{Odd Parcel.}
A set of visually identical sealed parcels contains one or two parcels that
differ in weight. The only instrument is a two-pan balance whose pans hold
several parcels each. The goal is to put every odd parcel, and nothing else, in the answer box.
Group weighing identifies it in few comparisons, while one-by-one weighing is
legal but slower.
\par\noindent\emph{Main parameters:} 4, 6, or 8 parcels and occlusion Visible, Look, or Uncover.
\par\noindent\emph{Goal text:} ``There are $\langle$\emph{parcels}$\rangle$ on the counters in this kitchen. Most of these parcels weigh the same; fewer than half weigh differently. Put every parcel that weighs differently in the box and nothing else, leave the balance where it stands, then press \texttt{Submit}. Other objects on the counter are not part of the task; you may move them if needed.''

\paragraph{Puzzle Box.}
A container is held closed by a chain of interlocking parts, where each part
blocks another until it is moved. All parts are visible with large handles,
but the blocking relations are not. The goal is to place the item inside on
the tray. The robot must discover the release order by trying moves and
observing what moves, and some wrong moves jam another part until reversed.
\par\noindent\emph{Main parameters:} chain length 4, 5, or 6 and cover absent or present.
\par\noindent\emph{Goal text:} ``Open the wooden puzzle box on the counter, take out the $\langle$\emph{item}$\rangle$ inside, and stand it inside the tray. The lid slides sideways but is locked by sliding bolts with round black handles: a part moves only when nothing blocks it, so work out the order. Any order that works is fine. Once the $\langle$\emph{item}$\rangle$ is inside the tray and you have let go of everything, press the red \texttt{SUBMIT} button. Your first submission ends the episode; a wrong submission or no submission fails.''

\paragraph{Wobbly Stand.}
A stand or table has an uneven support, tilting its surface by an amount too
small to see directly. A ball placed on the surface rolls toward the low side
and reveals the tilt. Shim blocks of different thicknesses are available. The
goal is a level surface on which the ball stays put, with everything released.
\par\noindent\emph{Main parameters:} 1 or 2 short legs and shim choice absent or present.
\par\noindent\emph{Goal text:} ``The stand is not level. Use the shims to level it so the ball stays still on top, then press \texttt{Submit}.''

\section{Fine-Grained Capabilities for \dataset{}}
\label{app:capa}

\Cref{tab:capability_matrix} lists the capabilities each task targets. The marks list the capabilities an efficient solution uses. Success alone does not prove them, because exhaustive strategies such as opening every compartment or trying every part in turn remain legitimate. Directed search is the core of the Search family, active perception of the Inspect family, and experimental identification of the Test family. Evidence integration and memory are needed wherever an observation only matters together with earlier ones. Affordance discovery is targeted in Unfamiliar Containers, where the hidden property is how each box opens. Physical causal inference is targeted in Puzzle Box and Wobbly Stand, where the agent learns the hidden structure from how the scene responds to its actions.

Consequential action is marked where gathering information can damage the goal, for example spilling a mug's contents or locking a token away. Long-horizon planning is marked where the steps must be ordered in advance, as in Locked Storage and Puzzle Box. Every task ends with a \texttt{SUBMIT} press, so deciding when the evidence is enough to commit is common to all tasks. It is not counted as consequential action in the table.

\begin{table}[h]
\centering
\footnotesize
\caption{\textbf{Capabilities each task targets,} with what each task hides ($\checkmark$: directly targeted, $\sim$: supporting, --: none). DS: directed search. AP: active perception. EI: evidence integration. M: memory. AD: affordance discovery. XI: experimental identification. CI: physical causal inference. CA: consequential action, marked where an information-gathering step can irreversibly damage the goal. LP: long-horizon planning. The table lists what each task demands, not where agents fail.}
\label{tab:capability_matrix}
\setlength{\tabcolsep}{3.5pt}
\begin{tabular}{@{}lllccccccccc@{}}
\toprule
& & & \multicolumn{2}{c}{\textcolor[HTML]{365F79}{\shortstack{Evidence\\acquisition}}} & \multicolumn{2}{c}{\textcolor[HTML]{526B48}{\shortstack{Evidence\\use}}} & \multicolumn{3}{c}{\textcolor[HTML]{65527F}{\shortstack{Interactive\\inference}}} & \multicolumn{2}{c}{\textcolor[HTML]{C7752B}{\shortstack{Action\\organization}}} \\
\cmidrule(lr){4-5}\cmidrule(lr){6-7}\cmidrule(lr){8-10}\cmidrule(lr){11-12}
Family & Task & Hidden information & DS & AP & EI & M & AD & XI & CI & CA & LP \\
\midrule
\multirow{3}{*}{Search}
 & Locked Storage & target behind locks & $\checkmark$ & -- & $\checkmark$ & $\sim$ & -- & -- & $\sim$ & $\checkmark$ & $\checkmark$ \\
 & Search Room & where the targets are & $\checkmark$ & $\sim$ & $\checkmark$ & $\checkmark$ & -- & -- & -- & -- & $\sim$ \\
 & Blackout Search & targets in the dark & $\checkmark$ & $\checkmark$ & $\checkmark$ & $\checkmark$ & -- & -- & -- & -- & $\sim$ \\
\midrule
\multirow{3}{*}{Inspect}
 & Painted Cubes & marks on unseen faces & -- & $\checkmark$ & $\checkmark$ & $\sim$ & -- & -- & -- & $\sim$ & $\sim$ \\
 & Marked Mugs & labels under vessels & -- & $\checkmark$ & $\sim$ & $\sim$ & -- & -- & -- & $\checkmark$ & $\sim$ \\
 & Unfamiliar Containers & how containers open & $\sim$ & -- & -- & $\sim$ & $\checkmark$ & $\sim$ & -- & -- & $\sim$ \\
\midrule
\multirow{4}{*}{Test}
 & Puzzle Box & bolt blocking order & -- & -- & -- & $\sim$ & $\sim$ & $\sim$ & $\checkmark$ & $\sim$ & $\checkmark$ \\
 & Stamp Composition & patterns and rotations & -- & -- & $\checkmark$ & $\sim$ & -- & $\checkmark$ & -- & $\checkmark$ & $\checkmark$ \\
 & Wobbly Stand & short legs and gaps & -- & -- & $\sim$ & $\sim$ & -- & $\checkmark$ & $\checkmark$ & $\checkmark$ & $\sim$ \\
 & Odd Parcel & which parcel is odd & -- & -- & $\checkmark$ & $\checkmark$ & -- & $\checkmark$ & -- & -- & $\sim$ \\
\bottomrule
\end{tabular}
\end{table}

\section{Task Progress}
\label{app:progress}

This appendix defines the progress $P\in[0,1]$ of every task. $P$ is computed on the scene frozen at the first \texttt{SUBMIT} press, a stop, or the end of the budget.

\paragraph{Search Room.} Each target earns a third of its credit once its compartment is open, two thirds once it has been taken out, and full credit once it stands upright on the tray, released. A target under a cloche or on the open counter earns nothing for opening. Credit for opening and taking out is kept even if the agent later closes the compartment or drops the target. $P$ is the mean credit over the targets, less 0.1 if a non-target object is on the tray. A compartment counts as open at 15\% of its joint range, which is 0.15\,m for a drawer and 0.24\,rad for a door. A target counts as taken out once it leaves its compartment or cloche, or once it moves 0.15\,m from an open spot.

\paragraph{Locked Storage.} The locks of the chain and the target share $P$ equally. Each lock that is open at the end earns its share, and the target earns its share as in Search Room. Once the target has been taken out, every lock counts as open. The dead-end compartment is not part of the chain. $P$ is reduced by 0.1 if a non-target object is on the tray.

\paragraph{Blackout Search.} Same as Search Room.

\paragraph{Marked Mugs.} $P=\frac{1}{n}\sum_i c_i\,b_i$. Let $f_i$ be the fraction of the footprint disc of vessel $i$ (radius 3.95\,cm) that lies on the pad of its label colour. The placement credit $c_i$ is 1 if $f_i\ge 3/4$, 0.5 if $1/2\le f_i<3/4$, and 0 otherwise, and it counts only while the vessel is upright, released, still and supported by that pad. A vessel on a wrong pad scores 0. $b_i=1$ if the vessel holds its own two balls, else 0.5. Success requires $c_i=b_i=1$ for every vessel.

\paragraph{Painted Cubes.} $P=c/n$, where $c$ is the number of cubes captured in their correct bin. Captures are irreversible.

\paragraph{Unfamiliar Containers.} As Search Room, over the items. An item earns a third of its credit once its box has been opened, two thirds once it is out of the box, and full credit once it is in the bowl.

\paragraph{Stamp Composition.} $P=\sum_{c\in C} g_c/(|C|+m)$, where $C$ is the set of target cells. A cell scores $g_c=1$ if it holds only clean dots, each centred within 9\,mm, and at most its $k_c$ dots. It scores $g_c=0.5$ if a dot crosses the cell edge or it holds more than $k_c$ dots, and $g_c=0$ if it is empty. $m$ counts the stray dots outside $C$, plus one if the ink outside the grid exceeds 35 pixels. Dots are located from the recorded impressions with the die geometry. Success also requires every stamp released, still and supported. An untouched scene scores 0.

\paragraph{Odd Parcel.} $P=\max\bigl(0,\,|B\cap O|/|B\cup O|-0.1\,[\text{balance moved}]\bigr)$, where $O$ is the set of odd parcels and $B$ the set of parcels released in the answer box. An empty box scores 0, and parcels left on the pans are not penalized. Moving the balance more than 2\,cm fails the episode.

\paragraph{Puzzle Box.} The bolts of the chain and the item share $P$ equally. Each bolt that is released at the end earns its share, and decoy bolts earn nothing. The item earns a third of its share once the lid is past half its travel, two thirds once the item has been out of the box, and its full share once it rests in the tray, released. Once the item has been out, every bolt counts as released. Success also requires that the robot touches no part at the press.

\paragraph{Wobbly Stand.} $P=q\,(1+b)/2$, where $b=1$ if the ball is on top of the stand and 0 otherwise. With $\theta_0$ and $\theta$ the initial and final tilt, $q=\operatorname{clip}(1-\theta/\theta_0,0,1)$ while the stand is released, settled and untouched, and $q=1$ within the level threshold. $q=0$ while the robot touches the stand, and values below 0.01 read 0. A level stand without the ball scores 0.5, and an untouched scene scores 0.

\begin{table}[h]
\centering
\small
\caption{\textbf{How episodes end,} in \% of 500 episodes per model. The first five columns sum to 100. Only a correct submission succeeds. Precision is the share of submissions that were correct.}
\label{tab:outcomes}
\setlength{\tabcolsep}{5pt}
\begin{tabular}{lrrrrrr}
\toprule
Model & \makecell{Correct\\submission} & \makecell{Wrong\\submission} & Stopped & \makecell{Reply with\\no action} & \makecell{Out of\\decisions} & Precision \\
\midrule
Gemini 4 Argon   & 23.8 & 30.2 & 12.8 & 0.2 & 33.0 & 44.1 \\
GPT-6 Astra      & 23.2 & 61.8 & 5.4  & 0.0 & 9.6  & 27.3 \\
Claude Opus 5.5  & 13.8 & 31.6 & 48.6 & 4.8 & 1.2  & 30.4 \\
GPT-6.1 Sol      & 12.2 & 62.8 & 3.4  & 0.0 & 21.6 & 16.3 \\
Claude Fable 5.1 & 11.4 & 53.6 & 28.6 & 5.4 & 1.0  & 17.5 \\
Gemini 3.8 Flash & 2.0  & 55.6 & 0.8  & 0.0 & 41.6 & 3.5  \\
\bottomrule
\end{tabular}
\end{table}

\FloatBarrier
\section{Execution Skill Test Details}
\label{app:control}

Every skill-test scene keeps the kitchen of an evaluation instance and holds only one object and its target from that task, with no \texttt{SUBMIT} button. The simulator checks the test's goal after every physics tick and ends the episode once the goal has held for 0.5\,s. Otherwise the episode ends after 50 decisions. Success is scored from the recorded episode by the source task's own rule for that stage. The scenes vary what the skill depends on, such as straight and detour routes for driving or the eleven opening mechanisms of the containers. Every model plays the same 20 scenes per skill.

\paragraph{Skill selection.} Every task ends by placing objects on or into a target on the counter, such as a tray or a bin. Most tasks also require driving the base to an object, the Search tasks most of all. Pick-and-place and driving are therefore the two general skills. The drawer and cabinet tests use the storage layout of the Search tasks. Marked Mugs, Odd Parcel, Stamp Composition and Puzzle Box mostly involve moving parts on the counter, which the general skills cover, so they have no test of their own.

\paragraph{Reaching the target.} \Cref{tab:reached} compares how often the robot reached the target of a skill with how often it completed the skill.

\begin{table}[h]
\centering
\small
\caption{\textbf{Reaching the target and completing the skill,} in \% of the 20 episodes per skill, \gpts{} / \opus{} / \sol{}.}
\label{tab:reached}
\setlength{\tabcolsep}{5pt}
\begin{tabular}{llcc}
\toprule
Skill & Reached when & Reached & Completed \\
\midrule
Open drawer/cabinet & the gripper touched the handle & 85 / 85 / 75 & 50 / 60 / 30 \\
Pick from drawer/cabinet & the gripper touched the object & 90 / 80 / 100 & 70 / 60 / 60 \\
Shim the short leg & the shim came within 3\,cm of the leg & 100 / 90 / 95 & 35 / 60 / 40 \\
Drive and pick & the base came within 0.8\,m of the object & 100 / 100 / 95 & 100 / 100 / 90 \\
\bottomrule
\end{tabular}
\end{table}

\paragraph{Skills in the full tasks.} \Cref{tab:skills_in_context} gives the two execution stages of retrieving a hidden target in the full tasks, beside the isolated test of the same skill (\cref{sec:findings}). A compartment counts as tried when the robot has touched or when the gripper closed within 12,cm of its handle. A grasp counts as tried when the robot has touched the target, or when the gripper closed within 12,cm of it. Decisions per opening are averaged over successful openings, from the last decision with the robot base at least 0.2\,m from the compartment's standing position, where the isolated test starts. Decisions are compared only for opening, because there the two settings start alike. Taking a target out in the full tasks starts from however far the model opened the compartment and includes finding the target. Placing a shim in the full tasks includes finding the short leg and the right shim, while the isolated test names the leg and provides a single shim. For boxes, the isolated test names the opening mechanism, while in the full task the model must find it by trying.

\begin{table}[h]
\centering
\caption{\textbf{Skills in the full tasks.} Opening the compartment or box of a hidden target once it was tried, and taking the target out once a grasp was tried. The last column gives the isolated test of the same skill (success \%, \gpts{} / \opus{} / \sol{}).}
\label{tab:skills_in_context}
\resizebox{\linewidth}{!}{%
\begin{tabular}{lrrrl}
\toprule
Stage & \gpts{} & \opus{} & \sol{} & Isolated test \\
\midrule
\multicolumn{5}{l}{\emph{Search Room and Locked Storage: target in a compartment}} \\
\midrule
Opened $|$ tried & 52/58 (90\%) & 28/36 (78\%) & 24/32 (75\%) & Open drawer/cabinet: 50 / 60 / 30 \\
Decisions per opening (mean) & 10.4 & 25.7 & 15.2 & Open drawer/cabinet: 22.1 / 30.8 / 35.8 \\
Taken out $|$ grasp tried & 30/38 (79\%) & 17/18 (94\%) & 13/16 (81\%) & Pick from drawer/cabinet: 70 / 60 / 60 \\
\midrule
\multicolumn{5}{l}{\emph{Unfamiliar Containers: item in a box}} \\
\midrule
Opened $|$ tried & 57/81 (70\%) & 35/77 (45\%) & 49/63 (78\%) & Open named container: 100 / 90 / 95 \\
Taken out $|$ grasp tried & 43/44 (98\%) & 24/26 (92\%) & 28/31 (90\%) & Pick from open container: 90 / 90 / 70 \\
\bottomrule
\end{tabular}}
\end{table}

\FloatBarrier
\paragraph{Open models on the full tasks.}\label{app:open_models} We ran Qwen3.8-27B and Qwen3.8-Flash on a subset chosen to give them the best chance, with the same interface, prompt and decision budget as the frontier models. We took the easiest task of each family, by mean success over the six frontier models: Puzzle Box (Test), Painted Cubes (Inspect) and Locked Storage (Search). Within each task we took the 20 of its 50 evaluation instances that the frontier models solved most often. \Cref{tab:open_models} compares both Qwen models with the frontier models' episodes on the same 60 instances from the main evaluation. Neither Qwen model solves any of them, and both make almost no progress. Their failures are not formatting errors, since over 99\% of their commands are valid tool calls. Together with the skill tests (\cref{tab:control}), this suggests they fail at basic control rather than only at exploration.

\begin{table}[h]
\centering
\small
\caption{\textbf{Open models on the full tasks.} Success rate (SR) and progress (Prog, \%) on the 20 instances per task that the frontier models solve most often.}
\label{tab:open_models}
\setlength{\tabcolsep}{4pt}
\begin{tabular}{l*{5}{rr}}
\toprule
& \multicolumn{2}{c}{\gpts{}} & \multicolumn{2}{c}{\opus{}} & \multicolumn{2}{c}{\sol{}} & \multicolumn{2}{c}{Qwen3.8-Flash} & \multicolumn{2}{c}{Qwen3.8-27B} \\
\cmidrule(lr){2-3}\cmidrule(lr){4-5}\cmidrule(lr){6-7}\cmidrule(lr){8-9}\cmidrule(lr){10-11}
Task & SR & Prog & SR & Prog & SR & Prog & SR & Prog & SR & Prog \\
\midrule
Puzzle Box (Test)       & 95 & 99.7 & 80 & 93.6 & 75 & 91.6 & 0 & 0.0 & 0 & 2.7 \\
Painted Cubes (Inspect) & 75 & 90.4 & 25 & 61.2 & 35 & 73.8 & 0 & 0.0 & 0 & 0.0 \\
Locked Storage (Search) & 70 & 89.5 &  5 & 46.5 & 15 & 45.1 & 0 & 1.2 & 0 & 0.0 \\
\bottomrule
\end{tabular}
\end{table}

\section{Demonstration Dataset Details}
\label{sec:demonstration_details}

\Cref{tab:dataset} lists the scripted information gathering of each oracle and the size of each task's dataset. The rest of this appendix gives the verification, the language annotations, the release format, and the conversion of the search demonstrations into tool calls.

\paragraph{Language annotations.} Every frame stores the three camera views, the robot state, the action, and two levels of language annotation. A \emph{stage} such as ``Open and look inside the nearer of the two boxes on your right'' contains finer \emph{subtasks} such as ``Try pulling the knob on the front of the box''. Objects are named only by their appearance and position as seen from the robot at that moment. Hidden information is never named before the robot has seen it. For example, an item in a closed box is not named until the box is open. Every control step belongs to exactly one subtask. Stages last 10 to 50\,s and subtasks about 10\,s, and \cref{tab:dataset} gives the mean number of subtasks per task.

\paragraph{Format.} The dataset is released in the LeRobot v2.1 format, one dataset per task. Each frame holds the three camera views as 256$\times$256 video at 20\,Hz, the 16-dimensional robot state and the 12-dimensional action. The goal text is the task string, and per-frame indices point into the stage and subtask tables. The release scripts regenerate the datasets at other resolutions from the recorded states.

\paragraph{Conversion to tool calls.} To check that the search tasks can be solved through the agents' interface, we convert each Search Room and Blackout Search demonstration into the agents' own arm, base and wait calls. The conversion replays the demonstration and replaces each oracle motion with the call an agent would issue for it. Each call is run in simulation from the demonstration's state, and the state it reaches is compared with the state the demonstration reaches at the same point. Consecutive base motions are merged into one call when the merged call still reaches that state. Search Room needs a median of 66 calls, and all 500 episodes stay within the 200-decision budget. Blackout Search needs a median of 122, and 96\% of its episodes stay within the budget. These counts are upper bounds, because the conversion follows the oracle's path, and an agent that plans its own path can use fewer.

\begin{table}[h]
\centering
\footnotesize
\caption{\textbf{Demonstration dataset.} Scripted information gathering in each oracle's route, which the oracle performs although it knows the hidden information. Every task has 500 successful, verified episodes. Minutes: mean episode length. Hours: interaction time at 20\,Hz. Subtasks: mean number of subtask annotations per episode.}
\label{tab:dataset}
\begin{tabular}{@{}l>{\raggedright\arraybackslash}p{5.9cm}rrr@{}}
\toprule
Task & Scripted information gathering & Minutes & Hours & Subtasks \\
\midrule
Locked Storage & follows the token chain & 3.7 & 31.1 & 16 \\
Search Room & opens compartments nearest-first until the targets are seen & 3.0 & 25.1 & 11 \\
Blackout Search & carries the lamp along a nearest-first search & 7.5 & 62.1 & 33 \\
\midrule
Painted Cubes & removes covers and turns each cube face by face & 5.1 & 42.4 & 14 \\
Marked Mugs & lifts and tilts each vessel to read its label & 3.4 & 28.0 & 15 \\
Unfamiliar Containers & visits boxes nearest-first and tries each knob's actions until one opens the box & 6.4 & 53.0 & 29 \\
\midrule
Puzzle Box & also pulls blocked bolts on the way & 1.6 & 13.7 & 10 \\
Stamp Composition & prints test impressions before the final one & 3.6 & 30.1 & 22 \\
Wobbly Stand & releases the ball, places a shim, and tests again & 2.8 & 23.2 & 11 \\
Odd Parcel & weighs parcels on the balance & 6.9 & 57.8 & 42 \\
\midrule
Total (10 $\times$ 500 episodes) & & 4.4 & 366.4 & 20 \\
\bottomrule
\end{tabular}
\end{table}
\FloatBarrier

\section{Workload and Occlusion Details}
\label{sec:factor_details}

\begin{figure}[t]
\centering
\includegraphics[width=0.85\linewidth]{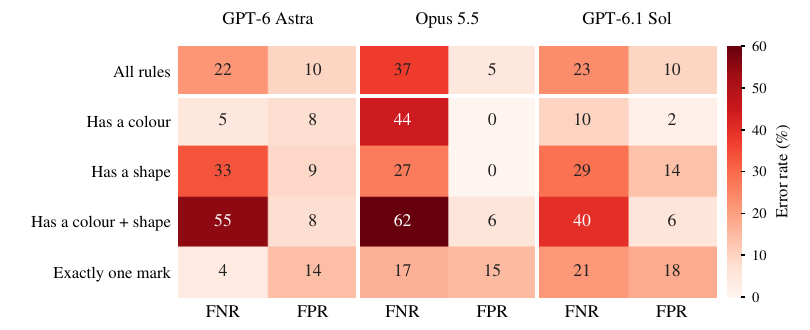}
\caption{\textbf{Painted Cubes: error rates by rule.} False-negative rate (FNR) and false-positive rate (FPR) over committed cubes, in \%.}
\label{fig:cubes_direction}
\end{figure}

This section reports how the two difficulty parameters affect the three models, and \cref{sec:findings} uses the occlusion result. Painted Cubes, Marked Mugs, Stamp Composition and Odd Parcel vary occlusion at three levels, and Unfamiliar Containers at the first two. In \emph{Visible}, every task object is in a camera view at the start. In \emph{Look}, some object starts out of view, for example behind a tall object, and the robot must move to see it. In \emph{Uncover}, some object is under a cover that the robot must remove, for example a cloche to lift or a board to push aside. Workload is the number of units a task requires, for example the number of cubes or the depth of the lock chain, and six tasks vary it at three levels. \Cref{tab:factors} gives success and progress per level and \cref{tab:factor_phases} the failure families per level. \Cref{fig:cubes_direction} gives the Painted Cubes error rates by rule.

\begin{table}[h]
\centering
\small
\caption{\textbf{Occlusion and workload.} Success and progress (\%) by level, pooled over the three models and per model. Occlusion on the four tasks that vary it, workload on the six.}
\label{tab:factors}
\setlength{\tabcolsep}{5pt}
\begin{tabular}{lrrrrrrrr}
\toprule
& \multicolumn{4}{c}{SR} & \multicolumn{4}{c}{Prog} \\
\cmidrule(lr){2-5}\cmidrule(lr){6-9}
& \makecell{GPT-6\\Astra} & \makecell{Opus\\5.5} & \makecell{GPT-6.1\\Sol} & Pooled & \makecell{GPT-6\\Astra} & \makecell{Opus\\5.5} & \makecell{GPT-6.1\\Sol} & Pooled \\
\midrule
\multicolumn{9}{l}{\textit{Occlusion (4 tasks)}} \\
\quad Visible & 28.8 & 16.3 & 14.0 & 19.7 & 57.6 & 45.8 & 39.5 & 47.7 \\
\quad Look & 16.3 & 7.8 & 5.0 & 9.7 & 53.5 & 38.6 & 32.6 & 41.6 \\
\quad Uncover & 22.3 & 7.5 & 6.1 & 12.0 & 53.1 & 35.9 & 34.1 & 41.1 \\
\quad Look $-$ Visible & $-$12.5 & $-$8.6 & $-$9.0 & $-$10.0 & $-$4.1 & $-$7.2 & $-$7.0 & $-$6.1 \\
\quad Uncover $-$ Visible & $-$6.5 & $-$8.8 & $-$7.9 & $-$7.7 & $-$4.6 & $-$9.9 & $-$5.4 & $-$6.6 \\
\midrule
\multicolumn{9}{l}{\textit{Workload (6 tasks)}} \\
\quad Low & 35.6 & 16.1 & 17.2 & 23.0 & 58.0 & 43.6 & 39.2 & 47.0 \\
\quad Mid & 15.4 & 4.0 & 4.6 & 8.0 & 50.0 & 30.2 & 33.3 & 37.8 \\
\quad High & 10.5 & 4.1 & 1.1 & 5.2 & 43.1 & 32.9 & 29.0 & 35.0 \\
\quad High $-$ Low & $-$25.1 & $-$12.1 & $-$16.1 & $-$17.7 & $-$14.9 & $-$10.8 & $-$10.2 & $-$12.0 \\
\bottomrule
\end{tabular}
\end{table}

\begin{table}[h]
\centering
\small
\caption{\textbf{Where the extra failures come from.} Unmet requirements per episode by failure family (execution failures and side effects summed), pooled over GPT-6 Astra, Opus 5.5 and GPT-6.1 Sol. Occlusion on the four tasks that vary it, workload on the six.}
\label{tab:factor_phases}
\setlength{\tabcolsep}{5pt}
\begin{tabular}{lrrrr}
\toprule
& Missing & Wrong & \makecell{Execution\\+ side effect} & All \\
\midrule
\multicolumn{5}{l}{\textit{Occlusion (4 tasks)}} \\
\quad Visible & 0.62 & 1.29 & 0.67 & 2.58 \\
\quad Look & 0.89 & 1.33 & 0.53 & 2.74 \\
\quad Uncover & 0.94 & 1.30 & 0.69 & 2.94 \\
\quad Look $-$ Visible & 0.27 & 0.04 & $-$0.14 & 0.16 \\
\quad Uncover $-$ Visible & 0.32 & 0.01 & 0.02 & 0.36 \\
\midrule
\multicolumn{5}{l}{\textit{Workload (6 tasks)}} \\
\quad Low & 0.49 & 0.76 & 0.45 & 1.69 \\
\quad Mid & 0.79 & 1.26 & 0.68 & 2.73 \\
\quad High & 1.33 & 1.34 & 0.72 & 3.39 \\
\quad High $-$ Low & 0.85 & 0.58 & 0.27 & 1.70 \\
\bottomrule
\end{tabular}
\end{table}

\section{Failure Attribution Details}
\label{sec:attribution}
\paragraph{Requirements.} We attribute failures per requirement rather than per episode, as a single episode can fail in multiple ways. An episode's requirements contains both its goal objectives (such as each target that must reach the tray) and its task constraints (such as keeping other objects off the tray). Each unmet requirement serves as an individual unit of analysis in our attribution procedure (\cref{alg:attribution}).
\begin{algorithm}[t]
\caption{Failure attribution over the failed episodes of one agent.}
\label{alg:attribution}
\begin{algorithmic}[1]
\Require For each failed episode $e$: the private task spec, the simulator state at every 20\,Hz tick, the commands, and the images $I^e_1,\dots,I^e_{T_e}$ delivered to the model at its decision ticks
\Ensure Share of failures in each family
\State $w[\cdot] \gets 0$ \Comment{weight per family}
\ForAll{failed episodes $e$}
  \State $U_e \gets$ unmet requirements of $e$ in its final state, as defined per task
  \ForAll{units $r \in U_e$} \Comment{every unmet requirement, not only the first}
    \State $w[\Call{Label}{r}] \gets w[\Call{Label}{r}] + 1/|U_e|$ \Comment{each failed episode has total weight 1}
  \EndFor
\EndFor
\State \Return $w[\ell] / \sum_{\ell'} w[\ell']$ for each family $\ell$
\Statex
\Function{Label}{$r$}
  \State Follow the events that led to $r$'s final state, ignoring errors that were later repaired
  \ForAll{rules $\rho$ of $r$'s task, in their fixed order}
    \If{$\rho$ applies to $r$} \Return the family of $\rho$ \EndIf
  \EndFor
\EndFunction
\Statex
\Function{Seen}{$f, t$} \Comment{called by rules that ask whether the agent saw $f$}
  \ForAll{decision ticks $t' < t$}
    \State $P \gets$ pixels of $f$ in $I^e_{t'}$, from a segmentation rendered from the state at $t'$
    \State $b \gets$ mean brightness of $P$, read from the saved image $I^e_{t'}$
    \If{$|P|$ and $b$ reach the thresholds for $f$} \Return true \EndIf
  \EndFor
  \State \Return false
\EndFunction
\end{algorithmic}
\end{algorithm}

\paragraph{Units.} Each unmet requirement in the final simulator state of an episode constitutes one unit, such as a target left off the tray or an incorrect stamp cell. When a single error leaves multiple requirements unmet, each requirement is treated as a distinct unit and attributed independently. Across all tasks, this yields 931 units for \gpts{} from 384 failed episodes, 1{,}143 for \opus{} from 431, and 1{,}182 for \sol{} from 439.

\paragraph{Weighting.} Each failed episode carries equal weight, split evenly across its units. We compute each family's share as the weighted fraction of its units. We also tried another aggregation scheme by pooling all units and averaging the ten per-task shares. Under both schemes, missing evidence and wrong decisions remain the two largest families for every model. Together they account for 64--76\% of failures, and no family share shifts by more than 8 percentage points (\cref{tab:failure_per_task}).

\paragraph{Measurements.} Attribution relies on data logged during execution: the simulator state recorded at every 20\,Hz tick and the three camera views delivered to the model at each decision step. Replaying these simulator states reproduces the exact scene geometry, allowing us to render ground-truth segmentation masks for each delivered observation. These masks identify the visible pixels of each object and key features (e.g., cube markings or container labels), whose pixel brightness is measured directly from the delivered images. From the physical trajectory, we also extract interaction events, including grasps, drops, final resting poses, opened compartments, and stamped ink impressions.

\paragraph{Decision procedure.} \Cref{alg:attribution} outlines the decision procedure. For each unit, we trace the sequence of events and identify the first error that was not subsequently repaired. Each task defines an ordered list of deterministic rules. We evaluate these rules sequentially to assign the failure family to each unit based on the first rule that matches. A failed recovery attempt is classified as an execution failure, whereas an unnoticed error is attributed to a wrong decision if the deciding evidence was observed by the agent, and to missing evidence if it was not. The entire procedure is programmatic and deterministic, evaluated directly on recorded run logs.

\paragraph{Family boundaries.} Several criteria are used to resolve edge cases between failure families. For premature terminations (stopping or submitting early), the unit is attributed to missing evidence if the agent stopped before observing the deciding evidence, and to a wrong decision if it stopped after observing the evidence. A failed physical attempt to uncover evidence (such as opening a drawer insufficiently) is attributed to execution failure rather than missing evidence. Finally, exploratory actions that perturb the scene (such as tipping a mug or rolling a test ball) are valid, so the unit is classified under side effect only to damage left unrepaired. Whether such damage was an unforeseen consequence or the result of clumsy execution cannot be determined from the recorded data.

\paragraph{Thresholds and release.} To determine whether an object was observed by the agent (\Call{Seen}{} in \cref{alg:attribution}), we evaluate its visibility across the delivered camera views. The saved simulator state identifies which pixels belong to the object, and the delivered image gives their brightness. An object counts as observed when its pixel count and mean brightness both meet fixed thresholds, set to $\ge 150$ visible pixels in one delivered image with mean brightness $\ge 25$ ($\ge 3$ in Blackout Search). In Odd Parcel, a parcel counts as determined when every odd set that best fits the settled weighings the agent saw agrees on it. The remaining thresholds, the rules of every task and the per-unit labels are released with the benchmark.

\paragraph{Manual check.} Visibility and weighings are computed from the simulator state, so the one judgment left in a missing-evidence label is whether the agent tried to open the hiding place. The pipeline counts an attempt when the robot touches the hiding place or commands its gripper to within 10--12\,cm of its handle. A human annotator checked 50 randomly sampled units against videos of the agent's camera views, blind to the automatic label, and 46 match. Of the four mismatches, three met the threshold unintentionally while the robot was manipulating other objects. In the fourth, the robot tried to open a box, but its gripper never came within 12\,cm of it.


\begin{table}[h]
\centering
\small
\caption{\textbf{Failure attribution per task,} in units (unmet requirements of failed episodes). Ran out: units in the Missing or Wrong columns whose episode ended by the decision budget.}
\label{tab:failure_per_task}
\setlength{\tabcolsep}{4pt}
\resizebox{\linewidth}{!}{%
\begin{tabular}{llrrrrrrr}
\toprule
Task & Policy & Failed episodes & Units & Missing & Wrong & Side effect & Execution & Ran out \\
\midrule
Locked Storage & \gpts{} & 36 & 87 & 11 & 49 & 0 & 27 & 5 \\
 & \opus{} & 47 & 118 & 21 & 62 & 0 & 35 & 1 \\
 & \sol{} & 44 & 108 & 15 & 74 & 0 & 19 & 35 \\
\addlinespace
Search Room & \gpts{} & 50 & 121 & 63 & 25 & 0 & 33 & 30 \\
 & \opus{} & 50 & 118 & 77 & 19 & 0 & 22 & 0 \\
 & \sol{} & 50 & 132 & 81 & 18 & 0 & 33 & 56 \\
\addlinespace
Blackout Search & \gpts{} & 50 & 128 & 67 & 14 & 0 & 47 & 8 \\
 & \opus{} & 50 & 122 & 81 & 14 & 0 & 27 & 0 \\
 & \sol{} & 50 & 127 & 81 & 16 & 2 & 28 & 18 \\
\addlinespace
Painted Cubes & \gpts{} & 35 & 80 & 38 & 34 & 1 & 7 & 0 \\
 & \opus{} & 43 & 124 & 46 & 67 & 0 & 11 & 4 \\
 & \sol{} & 41 & 96 & 40 & 44 & 0 & 12 & 4 \\
\addlinespace
Marked Mugs & \gpts{} & 31 & 80 & 5 & 28 & 19 & 28 & 0 \\
 & \opus{} & 42 & 113 & 7 & 59 & 11 & 36 & 0 \\
 & \sol{} & 47 & 149 & 5 & 55 & 24 & 65 & 8 \\
\addlinespace
Unfamiliar Containers & \gpts{} & 47 & 111 & 70 & 13 & 0 & 28 & 15 \\
 & \opus{} & 50 & 126 & 79 & 11 & 0 & 36 & 0 \\
 & \sol{} & 50 & 125 & 83 & 15 & 0 & 27 & 38 \\
\addlinespace
Puzzle Box & \gpts{} & 2 & 2 & 1 & 0 & 0 & 1 & 0 \\
 & \opus{} & 16 & 26 & 1 & 22 & 0 & 3 & 0 \\
 & \sol{} & 14 & 17 & 3 & 10 & 0 & 4 & 3 \\
\addlinespace
Stamp Composition & \gpts{} & 43 & 195 & 45 & 102 & 17 & 31 & 0 \\
 & \opus{} & 49 & 260 & 9 & 193 & 15 & 43 & 0 \\
 & \sol{} & 46 & 280 & 85 & 167 & 16 & 12 & 4 \\
\addlinespace
Wobbly Stand & \gpts{} & 45 & 45 & 2 & 6 & 36 & 1 & 1 \\
 & \opus{} & 38 & 38 & 4 & 11 & 23 & 0 & 0 \\
 & \sol{} & 48 & 48 & 3 & 4 & 40 & 1 & 2 \\
\addlinespace
Odd Parcel & \gpts{} & 45 & 82 & 60 & 9 & 4 & 9 & 10 \\
 & \opus{} & 46 & 98 & 70 & 12 & 1 & 15 & 6 \\
 & \sol{} & 49 & 100 & 84 & 10 & 0 & 6 & 16 \\
\midrule
All tasks & \gpts{} & 384 & 931 & 362 & 280 & 77 & 212 & 69 \\
 & \opus{} & 431 & 1143 & 395 & 470 & 50 & 228 & 11 \\
 & \sol{} & 439 & 1182 & 480 & 413 & 82 & 207 & 184 \\
\bottomrule
\end{tabular}%
}
\end{table}

\paragraph{Breakdown by physical outcome.} \Cref{tab:end_commit} separates failures in the two failure families (Missing evidence and Wrong decision) by the agent's physical action. An object is considered as never placed if it remains in an unopened compartment or is put down short of its destination. An object is considered as placed wrongly if it is committed before deciding evidence is uncovered, or placed in contradiction to observed evidence.

\begin{table}[h]
\centering
\small
\caption{\textbf{Missing evidence and wrong decisions by what the agent did,} in \% of failures, each failed episode weighted equally.}
\label{tab:end_commit}
\setlength{\tabcolsep}{5pt}
\begin{tabular}{llrrr}
\toprule
Family & Object & \gpts{} & \opus{} & \sol{} \\
\midrule
Missing evidence & Never placed & 27.8 & 28.8 & 29.5 \\
 & Placed before the evidence decided & 11.2 & 9.4 & 11.4 \\
\addlinespace
Wrong decision & Never placed & 14.8 & 21.9 & 19.5 \\
 & Placed against the evidence & 10.2 & 11.9 & 9.3 \\
\midrule
Both & Never placed & 42.6 & 50.7 & 49.0 \\
 & Placed wrongly & 21.4 & 21.3 & 20.7 \\
\bottomrule
\end{tabular}
\end{table}

\FloatBarrier
\end{document}